%% file: main.tex
\documentclass{article}
\usepackage{iclr2027_conference,times}
\iclrfinalcopy
\input{preamble}

\input{numbers}

\input{numbers_v2}

\title{Texture Interpolation as Texton Matching}
\author{Ido Madar \\
Tel Aviv University \\
\texttt{idomadar@mail.tau.ac.il}
\And
Gal Oren \\
Stanford University \\
\texttt{galoren@stanford.edu}
\And
Shai Avidan \\
Tel Aviv University \\
\texttt{avidan@eng.tau.ac.il}
}
\begin{document}
\maketitle
\lhead{Preprint}
\input{sections/abstract}
\suppressfloats[t]
\input{figures/overview}
\input{sections/opening}
\input{sections/introduction}

\input{sections/related_work}
\input{sections/method}
\input{sections/couplings}
\input{sections/benchmark}
\input{sections/mechanism}
\input{sections/boundaries}

\input{sections/generative}
\input{sections/applications}
\input{sections/discussion}
\label{main:lastpage}
\clearpage
\input{sections/statements}
\bibliographystyle{iclr2027_conference}
\bibliography{references}
\clearpage
\appendix
\input{appendix/supplement}
\end{document}

%% file: preamble.tex
\usepackage[utf8]{inputenc}
\usepackage[T1]{fontenc}
\usepackage{amsmath,amssymb,mathtools}
\usepackage{graphicx,booktabs,multirow,makecell}
\usepackage{microtype,xspace,xcolor}
\usepackage{tikz,pgfplots}
\usetikzlibrary{arrows.meta,calc,positioning}
\pgfplotsset{compat=1.18}
\usepackage{caption,subcaption}
\usepackage{placeins}
\usepackage{hyperref}
\usepackage[capitalize,noabbrev]{cleveref}
\hypersetup{colorlinks=true,linkcolor=blue!55!black,citecolor=blue!55!black,
  urlcolor=blue!55!black,pdftitle={Texture Interpolation as Texton Matching},
  pdfauthor={Ido Madar, Gal Oren, Shai Avidan}}
\graphicspath{{figures/}}
\DeclareMathOperator*{\argmin}{arg\,min}
\DeclareMathOperator{\Cov}{Cov}

\newcommand{\etal}{et al.\xspace}
\definecolor{sourceblue}{RGB}{36,113,163}
\definecolor{targetorange}{RGB}{221,116,40}
\definecolor{transportgreen}{RGB}{41,137,77}
\definecolor{lightpanel}{RGB}{246,247,249}
\newif\ifstandalonesupp
\newcommand{\panelheader}[2]{\makebox[\dimexpr\linewidth/#1\relax][c]{\scriptsize #2}}
\newcommand{\sixpanelheaders}[6]{%
\panelheader{6}{#1}\panelheader{6}{#2}\panelheader{6}{#3}%
\panelheader{6}{#4}\panelheader{6}{#5}\panelheader{6}{#6}\par\nobreak}
\newcommand{\fivepanelheaders}[5]{%
\panelheader{5}{#1}\panelheader{5}{#2}\panelheader{5}{#3}%
\panelheader{5}{#4}\panelheader{5}{#5}\par\nobreak}
\newcommand{\fourpanelheaders}[4]{%
\panelheader{4}{#1}\panelheader{4}{#2}\panelheader{4}{#3}\panelheader{4}{#4}\par\nobreak}

%% file: numbers.tex
\newcommand{\FeatureDim}{384}
\newcommand{\NumSamples}{4096}
\newcommand{\SinkhornEpsilon}{0.05}
\newcommand{\SinkhornSteps}{100}
\newcommand{\NumGaussians}{101}

\newcommand{\FIDPixel}{108.0}
\newcommand{\KIDPixel}{0.030}
\newcommand{\IncPixel}{0.925}

\newcommand{\PlacePixel}{0.048}

\newcommand{\FIDGaussian}{97.9}
\newcommand{\KIDGaussian}{0.024}
\newcommand{\IncGaussian}{0.926}

\newcommand{\PlaceGaussian}{0.105}
\newcommand{\FIDTransport}{94.7}
\newcommand{\KIDTransport}{0.021}
\newcommand{\IncTransport}{0.942}

\newcommand{\PlaceTransport}{0.081}
\newcommand{\FIDHard}{112.8}
\newcommand{\KIDHard}{0.033}
\newcommand{\IncHard}{0.951}

\newcommand{\PlaceHard}{0.044}

\newcommand{\RuleKIDRandom}{0.077}
\newcommand{\RuleKIDMean}{0.025}
\newcommand{\RuleKIDPixel}{0.031}
\newcommand{\RuleKIDPosition}{0.043}
\newcommand{\RuleKIDGaussian}{0.022}
\newcommand{\RuleKIDNN}{0.027}
\newcommand{\RuleKIDHard}{0.030} %
\newcommand{\RuleKIDTransport}{0.020}
\newcommand{\RuleKIDSharp}{0.019}
\newcommand{\RuleFIDRandom}{224.3}
\newcommand{\RuleFIDMean}{183.7}
\newcommand{\RuleFIDPixel}{188.7}
\newcommand{\RuleFIDPosition}{211.8}
\newcommand{\RuleFIDGaussian}{177.6}
\newcommand{\RuleFIDNN}{182.4}
\newcommand{\RuleFIDHard}{184.2}
\newcommand{\RuleFIDTransport}{174.9}
\newcommand{\RuleFIDSharp}{173.7}
\newcommand{\RandomIncRealism}{1.035}
\newcommand{\RandomDINORealism}{0.926}

\newcommand{\EntropyCeiling}{8.32}
\newcommand{\TargetVariance}{38\%}

\newcommand{\TimeVacherPath}{576.6}

\newcommand{\DINOPlacementWins}{37}
\newcommand{\SpreadDINOLin}{60\%}
\newcommand{\SpreadDINOTransport}{66\%}
\newcommand{\SpreadSAMLin}{57\%}
\newcommand{\SpreadSAMTransport}{71\%}
\newcommand{\SpreadCNTElementLin}{60\%}
\newcommand{\SpreadCNTElementTransport}{87\%}
\newcommand{\SpreadCNTFieldLin}{67\%}
\newcommand{\SpreadCNTFieldTransport}{99\%}
\newcommand{\RoundTripTransport}{0.17}
\newcommand{\RoundTripFloor}{0.13}
\newcommand{\RoundTripPosition}{0.32}

\newcommand{\ReconReference}{2819}
\newcommand{\GaussianRidge}{10^{-4}}
\newcommand{\EpsLow}{0.005}

\newcommand{\EntropyLow}{7.09}
\newcommand{\EntropyHigh}{8.31}
\newcommand{\VarianceLow}{88\%}
\newcommand{\VarianceHigh}{1\%}

\newcommand{\HardEndpointGap}{0.80}

\newcommand{\RhoHealthy}{0.977\pm0.021}
\newcommand{\RhoFlagged}{0.52\pm0.08}

\newcommand{\EncodeSec}{0.09}\newcommand{\OTTransportSec}{0.03}\newcommand{\HardAssignSec}{0.48}\newcommand{\DecSec}{0.08}\newcommand{\DecSecLow}{0.02}
\newcommand{\OTsec}{0.21}\newcommand{\PathSecOT}{1.02}
\newcommand{\HardSec}{0.66}\newcommand{\LinearSec}{0.18}\newcommand{\PathSecLinear}{0.99}
\newcommand{\TMsec}{0.012}\newcommand{\TMEncSec}{0.009}\newcommand{\TMInterpSec}{0.003}

\newcommand{\TMVarRetOt}{0.33}\newcommand{\TMEpsRel}{0.03}

\newcommand{\EqGaphard}{0.81}\newcommand{\EqGapotSharp}{0.84}\newcommand{\EqGapot}{0.87}
\newcommand{\EqVarRetSharp}{0.88}\newcommand{\EqVarRet}{0.38}\newcommand{\EqBalP}{1.7\times10^{-14}}\newcommand{\EqGapP}{9.5\times10^{-13}}
\newcommand{\AlignLinZero}{0}\newcommand{\AlignLinSixteen}{15.2}\newcommand{\AlignLinMax}{23.9}\newcommand{\AlignGaussDev}{18.2}\newcommand{\AlignOTdev}{22.4}\newcommand{\AlignCrossLo}{16}\newcommand{\AlignCrossHi}{64}\newcommand{\AlignN}{40}
\newcommand{\GEdKIDmorph}{+0.0433}\newcommand{\GEdKIDmorphCI}{[+0.0333,+0.0534]}
\newcommand{\GEdKIDfix}{+0.0001}\newcommand{\GEdKIDfixCI}{[-0.0041,+0.0043]}
\newcommand{\GEScrambleDReal}{0.001}\newcommand{\GEKIDprimHard}{0.030}
\newcommand{\CeilKIDorig}{0.009}\newcommand{\CeilKIDrec}{0.014}\newcommand{\CeilKIDrecTwo}{0.025}\newcommand{\CeilKIDmid}{0.023}\newcommand{\CeilLPIPSrec}{0.21}\newcommand{\CeilRealOrig}{1.00}\newcommand{\CeilRealRec}{0.90}\newcommand{\CeilRealMid}{0.85}\newcommand{\CeilRho}{0.08}
\newcommand{\CNExact}{1{,}024}\newcommand{\SubsampleM}{4{,}096}\newcommand{\VacherWSec}{52.4}
\newcommand{\GEPrimN}{30}\newcommand{\SinkhornMargTol}{10^{-3}}

\newcommand{\GFIDpix}{218.9}\newcommand{\GRIncpix}{0.913}\newcommand{\GRDinpix}{0.856}\newcommand{\GPlacepix}{0.041}
\newcommand{\GFIDtex}{247.9}\newcommand{\GRInctex}{0.897}\newcommand{\GRDintex}{0.849}\newcommand{\GPlacetex}{0.040}
\newcommand{\GFIDgauss}{216.2}\newcommand{\GRIncgauss}{0.919}\newcommand{\GRDingauss}{0.847}\newcommand{\GPlacegauss}{0.103}
\newcommand{\GFIDot}{208.2}\newcommand{\GRIncot}{0.924}\newcommand{\GRDinot}{0.851}\newcommand{\GPlaceot}{0.081}
\newcommand{\GFIDhard}{211.8}\newcommand{\GRInchard}{0.943}\newcommand{\GRDinhard}{0.871}\newcommand{\GPlacehard}{0.044}
\newcommand{\GFIDtm}{204.0}\newcommand{\GRInctm}{0.948}\newcommand{\GRDintm}{0.907}\newcommand{\GPlacetm}{0.061}
\newcommand{\GFIDgpt}{193.7}\newcommand{\GRIncgpt}{0.960}\newcommand{\GRDingpt}{0.906}\newcommand{\GPlacegpt}{0.054}
\newcommand{\GFIDgptsym}{201.1}\newcommand{\GRIncgptsym}{0.953}\newcommand{\GRDingptsym}{0.969}\newcommand{\GPlacegptsym}{0.046}

\newcommand{\TMFIDot}{204.0}\newcommand{\TMRIncot}{0.942}\newcommand{\TMRDinot}{0.922}\newcommand{\TMBalot}{0.072}

\newcommand{\EqFIDlinear}{189.3}\newcommand{\EqBallinear}{0.046}\newcommand{\EqRInclinear}{0.923}
\newcommand{\EqFIDhard}{183.4}\newcommand{\EqBalhard}{0.043}\newcommand{\EqRInchard}{0.952}
\newcommand{\EqFIDotSharp}{174.5}\newcommand{\EqBalotSharp}{0.067}\newcommand{\EqRIncotSharp}{0.933}
\newcommand{\EqFIDot}{177.6}\newcommand{\EqBalot}{0.072}\newcommand{\EqRIncot}{0.931}
\newcommand{\EqFIDpaper}{174.9}\newcommand{\EqBalpaper}{0.076}\newcommand{\EqRIncpaper}{0.942}
\newcommand{\EqFIDgainOT}{8\%}
\newcommand{\EqdKIDotVsLinear}{-0.0107}\newcommand{\EqdKIDotVsLinearCI}{[-0.0154,-0.0059]}
\newcommand{\EqdKIDotPaperVsLinear}{-0.0124}\newcommand{\EqdKIDotPaperVsLinearCI}{[-0.0172,-0.0076]}
\newcommand{\EqdKIDhardVsLinear}{-0.0022}\newcommand{\EqdKIDhardVsLinearCI}{[-0.0066,+0.0022]}
\newcommand{\EqdKIDhard}{+0.0085}\newcommand{\EqdKIDhardCI}{[+0.0049,+0.0120]}
\newcommand{\EqdKIDhardSharp}{+0.0103}\newcommand{\EqdKIDhardSharpCI}{[+0.0070,+0.0135]}
\newcommand{\EqdKIDsharp}{-0.0018}\newcommand{\EqdKIDsharpCI}{[-0.0035,-0.0001]}

\newcommand{\TMdKIDhard}{+0.0087}\newcommand{\TMdKIDhardCI}{[+0.0042,+0.0132]}

\newcommand{\GdKIDotVsPix}{-0.0114}\newcommand{\GdKIDotVsPixCI}{[-0.0174,-0.0054]}

\newcommand{\GKIDpix}{0.0305}
\newcommand{\GKIDtex}{0.0504}
\newcommand{\GKIDgauss}{0.0253}
\newcommand{\GKIDot}{0.0190}
\newcommand{\GKIDhard}{0.0302}
\newcommand{\GKIDtm}{0.0119}
\newcommand{\GKIDgpt}{0.0051}
\newcommand{\GKIDgptsym}{0.0087}

\newcommand{\TMKIDot}{0.0099}

\newcommand{\EqKIDlinear}{0.0322}
\newcommand{\EqKIDhard}{0.0300}
\newcommand{\EqKIDotSharp}{0.0197}
\newcommand{\EqKIDot}{0.0216}
\newcommand{\EqKIDpaper}{0.0198}
\newcommand{\EqKIDgainOT}{38\%}
\newcommand{\GEKIDfixP}{0.030}\newcommand{\GEKIDfixA}{0.030}\newcommand{\GEKIDfixPA}{0.027}\newcommand{\GEKIDmorphP}{0.036}\newcommand{\GEKIDmorphA}{0.080}
\newcommand{\GEKIDprimSoft}{0.016}\newcommand{\GEKIDdenseOT}{0.019}

\newcommand{\AnchorChangeA}{46\%}\newcommand{\AnchorChangeB}{38\%}

\newcommand{\TMkN}{1000}\newcommand{\TMkKIDlerp}{0.0121}\newcommand{\TMkKIDmean}{0.0090}\newcommand{\TMkKIDot}{0.0093}
\newcommand{\TMkRInclerp}{0.948}\newcommand{\TMkRIncmean}{0.970}\newcommand{\TMkRIncot}{0.955}\newcommand{\TMkBallerp}{0.056}\newcommand{\TMkBalmean}{0.077}\newcommand{\TMkBalot}{0.078}
\newcommand{\TMkdKIDmean}{-0.0031}\newcommand{\TMkdKIDmeanCI}{[-0.0044,-0.0018]}\newcommand{\TMkdKIDot}{-0.0028}\newcommand{\TMkdKIDotCI}{[-0.0039,-0.0017]}
\newcommand{\TMkdKIDotVsMean}{+0.0003}\newcommand{\TMkdKIDotVsMeanCI}{[-0.0006,+0.0012]}\newcommand{\TMkRealP}{0.005}\newcommand{\TMkKIDgain}{26\%}

\newcommand{\CKIDpctGauss}{20\%}\newcommand{\CKIDpctOT}{30\%}\newcommand{\CKIDpctExact}{$-$10\%}
\newcommand{\TMkKIDpctOT}{23\%}

\newcommand{\GPTsec}{38}
\newcommand{\TMkFIDlerp}{82.4}\newcommand{\TMkFIDmean}{79.6}\newcommand{\TMkFIDot}{79.1}
\newcommand{\RuleBalRandom}{0.035}\newcommand{\RuleGapRandom}{1.00}
\newcommand{\RuleBalMean}{0.082}\newcommand{\RuleGapMean}{0.91}\newcommand{\RuleIncMean}{0.921}
\newcommand{\RuleBalPixel}{0.044}\newcommand{\RuleGapPixel}{0.29}\newcommand{\RuleIncPixel}{0.923}
\newcommand{\RuleBalPosition}{0.035}\newcommand{\RuleGapPosition}{0.00}\newcommand{\RuleIncPosition}{0.895}
\newcommand{\RuleBalGaussian}{0.107}\newcommand{\RuleGapGaussian}{0.81}\newcommand{\RuleIncGaussian}{0.928}
\newcommand{\RuleBalNN}{0.046}\newcommand{\RuleGapNN}{0.81}\newcommand{\RuleIncNN}{0.951}
\newcommand{\RuleBalHard}{0.043}\newcommand{\RuleGapHard}{0.81}\newcommand{\RuleIncHard}{0.953}
\newcommand{\RuleBalTransport}{0.081}\newcommand{\RuleGapTransport}{0.87}\newcommand{\RuleIncTransport}{0.933}
\newcommand{\RuleBalSharp}{0.083}\newcommand{\RuleGapSharp}{0.84}\newcommand{\RuleIncSharp}{0.942}
\newcommand{\GEKIDmorphPA}{0.046}

%% file: numbers_v2.tex
\newcommand{\GPlaceMedgpt}{0.456}

\newcommand{\GgptDRDinMin}{+0.035}\newcommand{\GgptDRDinMax}{+0.074}\newcommand{\GgptDRDinTM}{-0.021}

\newcommand{\MeanRealismDelta}{-0.016}
\newcommand{\MeanRealismCI}{[-0.029,-0.003]}

%% file: sections/abstract.tex
\begin{abstract}
Texture interpolation asks what should lie between two textures. Standard methods typically blend pixels or features at corresponding spatial locations, implicitly assuming that co-located elements should be mixed. For unrelated textures, however, this correspondence is arbitrary: a crack at one location has no particular relationship to a fiber at the same location in another image.

We study whether this hidden pairing choice is itself a major determinant of interpolation quality. We represent textures as collections of local appearance vectors, or textons, and replace positional pairing with appearance-aware matching while keeping the encoder, representation, and decoder fixed. This isolates the effect of correspondence from changes in model capacity or training.

Across texture pairs, appearance-aware matching consistently improves the distributional quality of interpolated midpoints over positional feature blending. Soft matching produces the strongest midpoint statistics, whereas one-to-one assignments yield more balanced transitions and closer agreement with the target endpoint. The same intervention also improves a second pretrained texture interpolator without retraining.

These results identify correspondence as a first-class design choice in texture interpolation: meaningful transitions depend not only on how strongly two textures are mixed, but also on which local appearances are mixed with one another.
\end{abstract}

%% file: figures/overview.tex
\begin{figure}[t]
\centering
\includegraphics[width=\linewidth]{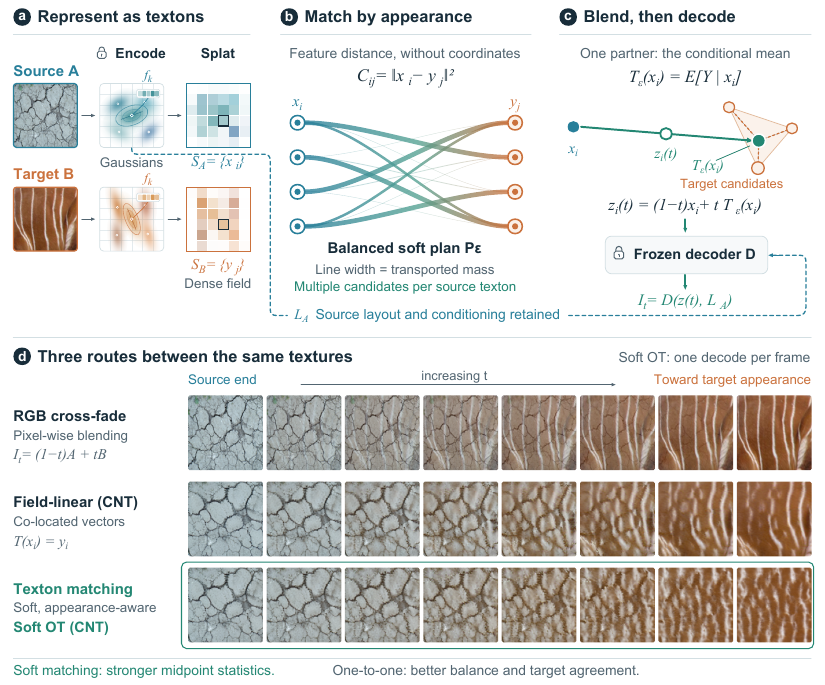}
\caption{\textbf{Texture interpolation is a choice of partners.}
Blending the same locations in unrelated textures can leave two patterns
overlaid. We instead associate local appearances, keeping the learned
representation and decoder fixed. (a) CNT splats localized Gaussian
primitives into dense fields of local appearance vectors (textons).
(b) Balanced entropic transport softly matches their vectors by appearance. (c) Each source vector moves
toward the weighted mean of its matches. One partner map is reused across
$t$, with source conditioning retained and one decoder pass per frame.
(d) Cracked earth to lined fur through RGB cross-fading, positional CNT
blending, and soft texton matching. The two CNT rows differ only
in pairing. Across experiments, soft matching improves midpoint statistics
without retraining, also in Texture Mixer, while one-to-one matching
improves midpoint balance and target agreement. Our prompted comparator
scores better at midpoints, but has no shared explicit path. The soft texton matching path is directed
and need not reconstruct the target. Panels (a--c) are schematic.}
\label{fig:overview}
\end{figure}

%% file: sections/opening.tex
\section{Introduction}
\label{sec:introduction}
What should lie between cracked earth and lined fur? A useful texture
transition forms a plausible material whose local appearance changes with the
interpolation parameter. An RGB cross-fade often leaves two transparent
patterns overlaid. Blending learned features can produce the same failure
(\cref{fig:overview}d). Both operations pair elements by spatial index, even
though a crack at one location has no special relationship to the fiber at
that location in an unrelated image.

%% file: sections/introduction.tex
Interpolation chooses a representation and an association between its
elements. Spatial pairing is useful for registered images. For unregistered
textures, appearance gives a more informative association. We test this
distinction by changing the partner map inside a fixed encoder and decoder
(\cref{fig:overview}a--c).

We demonstrate this idea on 
Compositional Neural Textures (CNT)~\citep{tu2024cnt}. CNT maps a texture image to a collection of textons, where each texton is represented with a Gaussian for its spatial location and an appearance vector. A texture image is rendered from the localized Gaussian primitives
with a learned decoder~\citep{tu2024cnt}. We align the dense appearance
vectors produced by splatting while retaining source conditioning:
soft texton matching (entropic OT) solves one balanced entropic transport
problem per pair, and each interpolation coefficient then costs one decoder
pass.

Gaussian alignment explains most of the improvement over positional blending.
Random permutation covers the target samples but ignores appearance, producing
typical-looking images with poor set-level statistics. These controls
measure the effect of matching without retraining a generator.

Similarly, Texture Mixer breaks a texture image into local features that capture  texton information. It then uses interpolation-specific training~\citep{yu2019texturemixer} for texture interpolation. This gives better interpolation statistics than texton matching. Yet, we show that replacing the positional latent blend of Texture Mixer with texton-matching lowers its KID
without retraining, while a one-to-one assignment raises it.

As an alternative, we evaluate GPT-Image-1.5 that synthesizes an intermediate image from the images and a description of the midway image. This gives better results than texton matching but is not parametric, does not form a controllable path, and comes at a higher computational price.

We make three contributions:
\begin{itemize}
\item A texton matching formulation and controlled comparison of texture
interpolation in a frozen CNT decoder, identifying appearance-aware
distribution alignment as the source of most of the improvement over
positional pairing, with an explicit account of barycentric contraction,
target-end behavior, and the cost of reusing a coupling across frames.
\item Evidence that the intervention transfers to a second pretrained
decoder (Texture Mixer), with the quantitative outcome reported in
\cref{sec:beyond}.
\item An evaluation of matching rules and learned generators that separates
midpoint statistics, endpoint balance, and path behavior, and isolates the
regime in which positional correspondence remains useful.
\end{itemize}

%% file: sections/related_work.tex
\section{Related Work}
\label{sec:related}
\paragraph{Textons.}
Julesz proposed that texture perception rests on local elements, textons,
and their statistics~\citep{julesz1981textons}. Parametric models made this
operational by matching joint statistics of wavelet
coefficients~\citep{portilla2000parametric}, and deep variants match the
statistics of convolutional features~\citep{gatys2015texture}. CNT makes the
elements explicit as localized Gaussian primitives with appearance
vectors~\citep{tu2024cnt}. Every such representation ends with a set of
local descriptors, which we call textons, so interpolation has to decide
which textons combine.
\paragraph{Texture transitions.}
Patch correspondence~\citep{ruiters2010patch}, image
melding~\citep{darabi2012melding}, stationary Gaussian
models~\citep{xia2014stationary}, and optimized neural
statistics~\citep{vacher2020texture} provide different routes to texture
mixing. Texture Mixer learns local and global latent spaces with
interpolation-specific objectives~\citep{yu2019texturemixer}. Mix-Max uses
content-derived priority maps to select among overlaid texels for spatial
material transitions~\citep{fournier2024mixmax}. Our study varies appearance
association inside an existing decoder, including rules that preserve
marginals and rules that do not.

\paragraph{Distribution matching.}
Statistical texture synthesis ranges from wavelet
models~\citep{portilla2000parametric} to deep feature
statistics~\citep{gatys2015texture}. Wasserstein texture
barycenters~\citep{rabin2012barycenter}, sliced Wasserstein
losses~\citep{heitz2021sliced}, and GOTEX~\citep{houdard2022gotex} already
use transport for synthesis or mixing. \citet{risser2020optimal} moves
autoencoder features toward their optimal-transport targets and interpolates
between them; we build on this operator and isolate the pairing rule as the
only variable inside a fixed encoder and decoder.
Gaussian feature transforms~\citep{li2017universal}, to which our Gaussian
row is related through covariance alignment (see also \citealp{huang2017adain}),
and transport on generative latents~\citep{agustsson2019latent} also
motivate our distributional controls.
We compare matching rules and their endpoint effects under a shared
representation and renderer, then evaluate the decoded results.
Full-reference texture metrics rest on the premise this paper builds on:
DISTS tolerates texture resampling by comparing spatial statistics of deep
features rather than co-located values~\citep{ding2022dists}, and the
information--estimation metric derives a distance from an image prior that
reduces to a second-moment distance for Gaussian priors~\citep{ohayon2026iem}.
Both find that low-order statistics of a good representation carry most of a
texture's perceptual content, which our Gaussian and target-mean rows show
from the operator's side; our object is an interpolation operator, not a
distance. We systematically vary the pairing rule within a shared frozen
encoder--decoder to separate the effects of appearance information, target
coverage, and matching sharpness; that isolation, not the transport
machinery, is the contribution.

\paragraph{CNT and correspondence.}
CNT matches latent Gaussians using centers and weights, then interpolates
their geometry and appearance; it explicitly proposes softer transport as a future
extension~\citep{tu2024cnt}. We study dense splat-field vectors, weighted by
their spatial coverage, rather than moving the sparse primitives. Exact OT (one-to-one) in our experiments acts on these same vectors. This separates the
object being matched from the hardness of its matching rule. Related transport
methods preserve content through self-similarity constraints in style
transfer~\citep{kolkin2019relaxed}.

%% file: sections/method.tex
\section{Interpolation with an Explicit Partner Map}
\label{sec:method}
Two routes lead from a pair of textures to a midpoint: a parametric operator
moves along a representation under an explicit coefficient $t$, and a
generative model describes the midpoint and samples it. This paper is about
the first; \cref{sec:generative} compares it with the second.
\subsection{From Gaussian primitives to a dense appearance field}
For an image $A$, CNT predicts about \NumGaussians{} localized Gaussian
primitives, of which about \GEPrimN{} carry most of the weight, with
appearance features $f_k$ and spatial parameters.
Splatting and alpha-compositing these primitives gives a conditioning field
$S_A(p)=\sum_k s_k(p)f_k$, where $s_k(p)$ is the contribution of primitive
$k$ at position $p$. The pretrained generator uses such fields at several
resolutions~\citep{tu2024cnt}. We operate on the finest field; each cell of
the field is a local appearance vector, a texton:
\begin{equation}
S_A=(x_i)_{i=1}^{N},\qquad S_B=(y_i)_{i=1}^{N},\qquad
N=256^2,\quad x_i,y_i\in\mathbb R^{\FeatureDim}.
\label{eq:fields}
\end{equation}
CNT's own interpolation matches and moves the primitives; we couple the
per-cell vectors beneath them and retain the source layout.
A field cell can mix several primitive features. Treating the cells as
uniformly weighted samples therefore weights appearance by spatial coverage;
it does not assign equal mass to each Gaussian. Coordinates are omitted from
the transport cost, but the splatted fields still contain spatial correlations.
Let $L_A$ denote the retained source layout and remaining decoder conditioning.

\subsection{One source-indexed path, several choices of partner}
Let $T$ assign a target appearance vector to each source texton. We define
\begin{equation}
z_i(t)=(1-t)x_i+tT(x_i),\qquad
I_t=D\big((z_i(t))_{i=1}^{N},L_A\big),\quad t\in[0,1].
\label{eq:path}
\end{equation}
The encoder and decoder $D$ are frozen (\cref{fig:overview}a--c). Positional
blending uses
$T(x_i)=y_i$; an appearance-based map instead associates nearby textons or
aligns the two distributions. The path keeps the partner map separate from
the representation and renderer, and admits Gaussian affine maps whose
outputs are not convex combinations of observed target samples.

\subsection{Balanced entropic transport}
For $M$ sampled vectors per field with cost $C_{ij}=\lVert x_i-y_j\rVert^2$
and uniform sampled measures, the balanced entropic objective is
\begin{equation}
P_\varepsilon=\argmin_{P\ge0}\langle P,C\rangle-\varepsilon H(P),
\quad P\mathbf1=P^\top\mathbf1=M^{-1}\mathbf1,
\quad H(P)=-\sum_{ij}P_{ij}\log P_{ij}.
\label{eq:transport}
\end{equation}
Sinkhorn iterations solve it~\citep{cuturi2013sinkhorn}. The barycentric
partner is
\begin{equation}
T_\varepsilon(x_i)=
\frac{\sum_jP_{\varepsilon,ij}y_j}{\sum_jP_{\varepsilon,ij}}
=M\sum_jP_{\varepsilon,ij}y_j.
\label{eq:barycentric}
\end{equation}
The normalized expression applies to the numerical plan; the last equality
holds for its prescribed marginals. Each unsampled cell uses the partner of
its nearest sampled source vector. We reuse this map for every $t$; we call
the resulting path soft texton matching; the unregularized assignment is
exact OT.

%% file: sections/couplings.tex
\subsection{What the matching controls}
\label{sec:couplings}
We use matching for any rule that pairs textons, including soft rules that
spread each texton over several counterparts; a coupling is the transport
plan that defines one. Field-linear pairs each cell with its co-located
texton, $T(x_i)=y_i$. Random permutation pairs through a random bijection,
$T(x_i)=y_{\sigma(i)}$, an appearance-blind negative control with exact
sample coverage. The target mean replaces every partner by $\bar y$.
Nearest-neighbor matching pairs each source texton with its nearest target
texton, $T(x_i)=y_{j^*(i)}$, and may reuse targets. The Gaussian $W_2$ rule
is the quadratic-cost optimal affine map between fitted Gaussians, ridge
$\GaussianRidge I$ (\cref{app:gaussian}). Entropic transport uses the plan
of \cref{eq:transport} and the conditional mean of \cref{eq:barycentric} on
\SubsampleM{} samples. Exact OT (one-to-one) is a Hungarian assignment under
squared feature cost on \CNExact{} samples, $T(x_i)=y_{\sigma^*(i)}$, with
the same nearest-neighbor extension.

The field-level controls share encoded inputs, source conditioning, and
decoder. The primitive-linear baseline instead interpolates CNT elements by
index before rendering, a representation-level reference distinct from the
native CNT application that first matches Gaussian primitives; RGB
cross-fading is an image-space reference.

\subsection{Projection, endpoints, and direction}
\label{sec:endpoints}
For a sampled pair $(X,Y)\sim P_\varepsilon$, the map satisfies
$T_\varepsilon(X)=\mathbb E[Y\mid X]$. Hence
\begin{equation}
\Cov(Y)=\Cov(T_\varepsilon(X))+\mathbb E[\Cov(Y\mid X)].
\label{eq:covariance}
\end{equation}
The second term is positive semidefinite. A diffuse plan therefore contracts
the projected target cloud; the measured default retains \TargetVariance{}
of target variance. The same relation holds along the path. If
$Z_t=(1-t)X+tY$ keeps the joint pairs and
$\bar Z_t=(1-t)X+t\mathbb E[Y\mid X]$ uses one partner per source, then
\begin{equation}
\Cov(Z_t)-\Cov(\bar Z_t)=t^2\mathbb E[\Cov(Y\mid X)].
\label{eq:path-contraction}
\end{equation}
Thus the distributional effect of projection grows toward the target end.
\Cref{app:projection} derives these identities and distinguishes the joint
displacement measure from the source-indexed field used by the decoder.

At $t=0$, \cref{eq:path} returns CNT's reconstruction of $A$. At $t=1$, it
decodes transported appearance with $L_A$. Exact OT (one-to-one) removes conditional variance on its sampled
cloud, but extension and source conditioning still affect the image endpoint;
the task is a directed transition, and the two directions can differ.

With an exact quadratic-cost plan the joint displacement path is a
$W_2$ geodesic~\citep{mccann1997convexity}; the projected, decoded path is
deterministic in $t$ and affine in the conditioning field once its samples
and plan are fixed, and its perceptual linearity remains empirical.

%% file: sections/benchmark.tex
\section{Experiments}
\label{sec:experiments}
\subsection{Protocol and primary comparison}
Unless otherwise stated, the default configuration uses \SubsampleM{}
sampled vectors per field,
\SinkhornSteps{} Sinkhorn iterations, $\varepsilon=\SinkhornEpsilon$ in raw
squared-feature units, and float32 arithmetic. The equal-sample and Texture
Mixer ablations iterate Sinkhorn to a marginal error below $\SinkhornMargTol$
and set $\varepsilon$ as stated in \cref{app:equal,app:tmcoupling}. The
primary comparison uses
1,000 stratified cross-category DTD
pairs~\citep{cimpoi2014dtd}. Endpoint images are stored at $128\times128$
(Texture Mixer's native resolution) and bicubically upsampled to
$256\times256$ before CNT encoding; all outputs are resized to $128$\,px
before feature extraction. FID/KID compare midpoint sets with
\ReconReference{} DTD references~\citep{heusel2017fid,binkowski2018kid}.
FID is the community standard but biased upward at small $n$; KID is
unbiased and admits a paired interval, so comparisons rest on KID.
The first 200 pairs form the matching ablation; a separate 120-pair subset
includes GPT-Image-1.5. Set-level scores are compared within each protocol.

We measure individual-sample manifold realism in Inception and DINOv2
features~\citep{kynkaanniemi2019precision,oquab2024dinov2}. Midpoint balance is
\begin{equation}
b(o;A,B)=\left|\frac{d(o,A)}{d(o,A)+d(o,B)}-\frac12\right|,
\label{eq:balance}
\end{equation}
using DINOv2 distance; balance measures relative proximity, FID/KID
agreement with the reference set. Path behavior is assessed separately.

\begin{table}[t]
\centering\small
\caption{\emph{Does replacing positional blending by texton matching improve
interpolation in a fixed decoder?} Primary comparison: within each block,
encoder, representation and decoder are fixed and the pairing of textons
changes; the rendering-controlled comparison, with paired intervals, is
\cref{tab:eqsample} (CNT, 200 pairs) and \cref{app:tmcoupling} (Texture
Mixer, 1,000 pairs); exact OT trades set-level statistics for balance. --:
not defined for the reference row.}
\label{tab:canonical}
\setlength{\tabcolsep}{5pt}
\begin{tabular}{lccccc}
\toprule
Method & KID $\downarrow$ & FID $\downarrow$ & Inc. real. $\uparrow$ & $b$ $\downarrow$ & $\Delta$KID vs.\ positional \\
\midrule
\multicolumn{6}{l}{\emph{CNT, 1,000 pairs}} \\
Positional blending (field-linear) & \KIDPixel & \FIDPixel & \IncPixel & \PlacePixel & -- \\
Texton matching: Gaussian moments & \KIDGaussian & \FIDGaussian & \IncGaussian & \PlaceGaussian & \CKIDpctGauss \\
Texton matching: Soft OT (ours) & \textbf{\KIDTransport} & \textbf{\FIDTransport} & \IncTransport & \PlaceTransport & \textbf{\CKIDpctOT} \\
Texton matching: Exact OT & \KIDHard & \FIDHard & \textbf{\IncHard} & \textbf{\PlaceHard} & \CKIDpctExact \\
\midrule
\multicolumn{6}{l}{\emph{Texture Mixer, 1,000 pairs}} \\
Positional blending (native) & \TMkKIDlerp & \TMkFIDlerp & \TMkRInclerp & \textbf{\TMkBallerp} & -- \\
Texton matching: Target mean & \textbf{\TMkKIDmean} & \TMkFIDmean & \textbf{\TMkRIncmean} & \TMkBalmean & \textbf{\TMkKIDgain} \\
Texton matching: Soft OT & \TMkKIDot & \textbf{\TMkFIDot} & \TMkRIncot & \TMkBalot & \TMkKIDpctOT \\
\bottomrule
\end{tabular}
\end{table}

Within the CNT decoder, soft texton matching improves FID/KID from \FIDPixel/\KIDPixel{}
for field-linear to \FIDTransport/\KIDTransport, Gaussian alignment is
close at \FIDGaussian/\KIDGaussian, and every soft texton-matching rule
beats positional blending on FID/KID within the same decoder, and exact OT
trades set-level statistics for balance and endpoint proximity. The gain
is distributional: positional blending has the lower measured balance error
($b=\PlacePixel{}$ against \PlaceTransport), and balance alone does not
distinguish a coherent intermediate from a superposition. Texture
Mixer, a separately trained
interpolator, is compared in \cref{sec:beyond}.

\begin{figure}[t]
\centering
\fivepanelheaders{$A$}{CNT-linear}{Gaussian $W_2$}{Soft OT (ours)}{$B$}
\includegraphics[width=\linewidth]{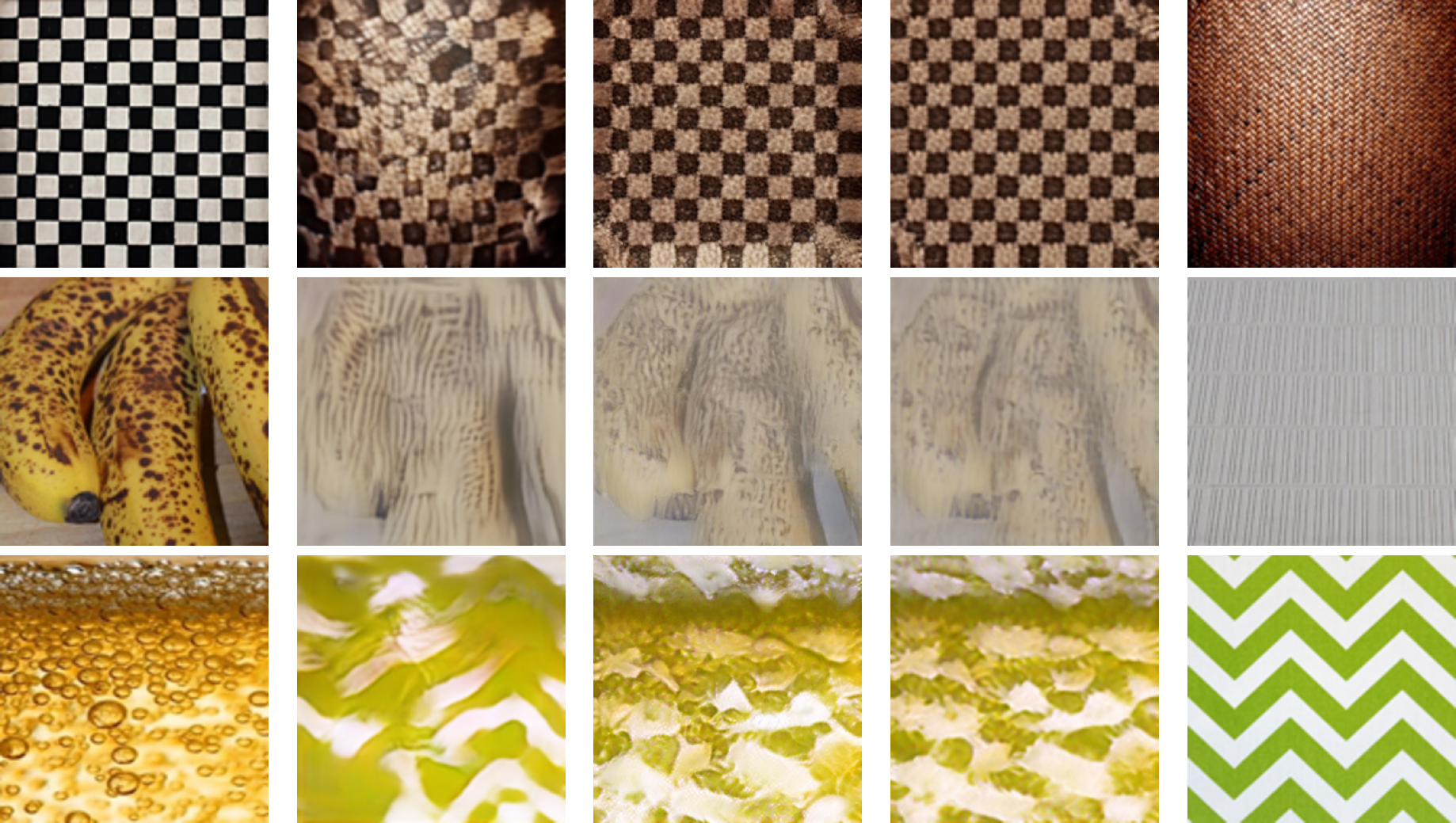}
\caption{Texton matching within the CNT decoder: rows 2, 4, and 5 of the
six-pair grid in \cref{app:qualitative}, Texture Mixer column omitted.
Columns: $A$, CNT-linear, Gaussian $W_2$, Soft OT (ours), $B$. Distribution
alignment blends appearance where the linear baseline retains endpoint
structures; Gaussian and empirical transport often look similar.}
\label{fig:qualitative}
\end{figure}

%% file: sections/mechanism.tex
\subsection{Which properties of the partner map matter?}
\label{sec:rules}
The 200-pair study varies appearance information, marginal constraints, and
assignment sharpness within the CNT encodings and renderer
(\cref{tab:rules}, \cref{app:rules}); it and the 1,000-pair benchmark use
the original rendering protocol, whereas every row of the matched-rendering
experiment (\cref{tab:eqsample}) shares the field blend and pyramid
construction.

\paragraph{Matched-rendering result.}
Under one rendering path and one scoring session (\cref{tab:eqsample}), the
operator lowers KID from \EqKIDlinear{} for positional pairing to
\EqKIDpaper{} (\EqdKIDotPaperVsLinear, CI \EqdKIDotPaperVsLinearCI;
\EqKIDgainOT{} lower KID, \EqFIDgainOT{} lower FID). A one-to-one assignment
does not show a statistically resolved improvement over positional pairing
(\EqdKIDhardVsLinear, CI \EqdKIDhardVsLinearCI) and loses \EqdKIDhard{} to
the soft plan (CI \EqdKIDhardCI); a sharp entropic plan is marginally better
than the default (\EqdKIDsharp, CI \EqdKIDsharpCI). Every soft
appearance-aware rule beats positional pairing on KID, and entropic
transport outperforms exact OT (one-to-one; cf.\ \citealp{ferradans2014regularized}),
so the form of the matching matters,
not appearance matching alone. The one-to-one assignment gives better
balance and a smaller target gap than the soft variants (\EqBalhard{}
against \EqBalot, $p=\EqBalP$; \EqGaphard{} against \EqGapot,
$p=\EqGapP$, both from the 200-pair matched-rendering study; field-linear's endpoint is the target field by construction;
gap values in \cref{app:equal}).
The regularization sweep is in \cref{fig:app-epsilon}.

\begin{table}[htbp]
\centering\small
\caption{\emph{Is the gain the matching itself, and which form of matching?}
One rendering path and one scoring session; the exact and \CNExact{}-sample
soft rows share the same sampled textons. Soft OT lowers KID by
\EqKIDgainOT{} against positional blending; exact OT does not show a
statistically resolved improvement over it.
Intervals and target-end gaps in \cref{app:equal}.}
\label{tab:eqsample}
\setlength{\tabcolsep}{5pt}
\begin{tabular}{lcccc}
\toprule
Rule & KID $\downarrow$ & FID $\downarrow$ & $b$ $\downarrow$ & Inc. real. $\uparrow$ \\
\midrule
Field-linear (same pipeline) & \EqKIDlinear & \EqFIDlinear & \EqBallinear & \EqRInclinear \\
Exact OT (\CNExact) & \EqKIDhard & \EqFIDhard & \textbf{\EqBalhard} & \textbf{\EqRInchard} \\
Soft OT \CNExact{} $\varepsilon=\EpsLow$ & \textbf{\EqKIDotSharp} & \textbf{\EqFIDotSharp} & \EqBalotSharp & \EqRIncotSharp \\
Soft OT \CNExact{} $\varepsilon=\SinkhornEpsilon$ & \EqKIDot & \EqFIDot & \EqBalot & \EqRIncot \\
Soft OT \SubsampleM{} (ours) & \EqKIDpaper & \EqFIDpaper & \EqBalpaper & \EqRIncpaper \\
\bottomrule
\end{tabular}
\end{table}

\paragraph{Sample coverage alone is insufficient.}
Random permutation has about four times the KID of either entropic plan
despite covering the same target samples, yet the highest Inception realism,
\RandomIncRealism{} (DINO \RandomDINORealism): individually typical outputs
whose ensemble poorly matches the reference.

\paragraph{Moment alignment accounts for most of the improvement.}
The Gaussian map reaches \RuleKIDGaussian{} KID against \RuleKIDTransport{}
for default entropic transport, with the same ordering on 1,000 pairs: the
larger change is from positional association to distribution alignment. A
target-mean map also attains low KID, \RuleKIDMean, with per-image realism
\MeanRealismDelta{} (CI \MeanRealismCI) relative to soft texton matching.

\paragraph{When does spatial correspondence help?}
\label{sec:boundary}
Across overlapping crops, controlled warps, and unrelated textures
(\cref{fig:app-boundary}), positional interpolation blurs or overlays the
endpoint patterns as alignment weakens. To separate alignment from content
we take \AlignN{} seamless images and let $B$ be a cyclic shift of $A$, so
the texture distribution is identical at every shift. Positional blending's
deviation from the source reconstruction grows monotonically with the shift
(\AlignLinZero{} at 0\,px, \AlignLinSixteen{} at 16, \AlignLinMax{} at 64,
DINOv2 distance), while the matching rules are far less sensitive to
displacement, staying at \AlignGaussDev--\AlignOTdev, an offset to which
projection, sampling, extension and rendering all contribute. On this
family the regimes cross between
\AlignCrossLo{} and \AlignCrossHi{}\,px (\cref{fig:app-alignshift}): a
reliable positional correspondence should be used when it exists, and texton
matching when it does not.

%% file: sections/boundaries.tex
\subsection{Does the effect extend beyond CNT?}
\label{sec:beyond}
\paragraph{The intervention transfers.}
Texture Mixer's representation and decoder are not ours. It blends its
$32\times32$ local latent positionally; replacing that pairing by texton
matching of the two latent clouds, rendered by its own generator with its
encodings and global-latent blend untouched, improves its set-level
statistics without retraining (\cref{tab:canonical}). The target-mean
matching lowers KID by \TMkdKIDmean{} (CI \TMkdKIDmeanCI, \TMkKIDgain) and
the soft matching by \TMkdKIDot{} (CI \TMkdKIDotCI) on \TMkN{} pairs, with
higher per-image realism ($p=\TMkRealP$); the two do not differ
(\TMkdKIDotVsMean, CI \TMkdKIDotVsMeanCI). One-to-one matching raises KID
(\TMdKIDhard, CI \TMdKIDhardCI, $n=120$) and gives the best balance. In this
latent the diffuse limit is the best set-level setting; the element-level
plan is what buys balance.

Appearance-aware matching also retains more source-feature spread than
positional blending in DINO and SAM~\citep{kirillov2023sam} feature spaces
(\cref{tab:app-representation}).

%% file: sections/generative.tex
\subsection{How does a prompted midpoint compare?}
\label{sec:generative}
We evaluate GPT-Image-1.5 on the same 120 DTD pairs through its
image-editing interface with both endpoints supplied. A directed prompt
requests the source arrangement with appearance halfway toward the target;
a symmetric prompt requests a natural midpoint. Each output is one
stochastic $1024^2$ image resized to $128^2$; intermediate coordinates are
separate calls.

\begin{table}[t]
\centering\small
\caption{\emph{How does texton matching compare with learned and prompted
generators, and at what cost?} Prompted generation gives the best midpoint
statistics at about \GPTsec{}\,s per output; texton matching needs no
training and pays its cost once per pair (\cref{app:runtime}).}
\label{tab:generative}
\setlength{\tabcolsep}{2pt}
\begin{tabular}{@{}lccccccl@{}}
\toprule
Method & FID $\downarrow$ & KID $\downarrow$ & Inc. real. $\uparrow$ & DINO real. $\uparrow$ & $b$ $\downarrow$ & s / midpoint & training \\
\midrule
Field-linear & \GFIDpix & \GKIDpix & \GRIncpix & \GRDinpix & \textbf{\GPlacepix} & \LinearSec & none \\
Soft OT (ours) & \GFIDot & \GKIDot & \GRIncot & \GRDinot & \GPlaceot & \OTsec & none \\
Exact OT & \GFIDhard & \GKIDhard & \GRInchard & \GRDinhard & \GPlacehard & \HardSec & none \\
Texture Mixer & \GFIDtm & \GKIDtm & \GRInctm & \GRDintm & \GPlacetm & \TMsec & GAN, domain-specific \\
Texture Mixer + soft OT & \TMFIDot & \TMKIDot & \TMRIncot & \TMRDinot & \TMBalot & \TMsec{}+\OTTransportSec & GAN, domain-specific \\
\midrule
GPT directed & \textbf{\GFIDgpt} & \textbf{\GKIDgpt} & \textbf{\GRIncgpt} & \GRDingpt & \GPlacegpt & \GPTsec{} (API) & proprietary \\
GPT symmetric & \GFIDgptsym & \GKIDgptsym & \GRIncgptsym & \textbf{\GRDingptsym} & \GPlacegptsym & \GPTsec{} (API) & proprietary \\
\bottomrule
\end{tabular}
\end{table}

\begin{figure}[t]
\centering
\sixpanelheaders{$A$}{Soft OT (ours)}{GPT directed}{GPT symmetric}{Texture Mixer}{$B$}
\includegraphics[width=\linewidth]{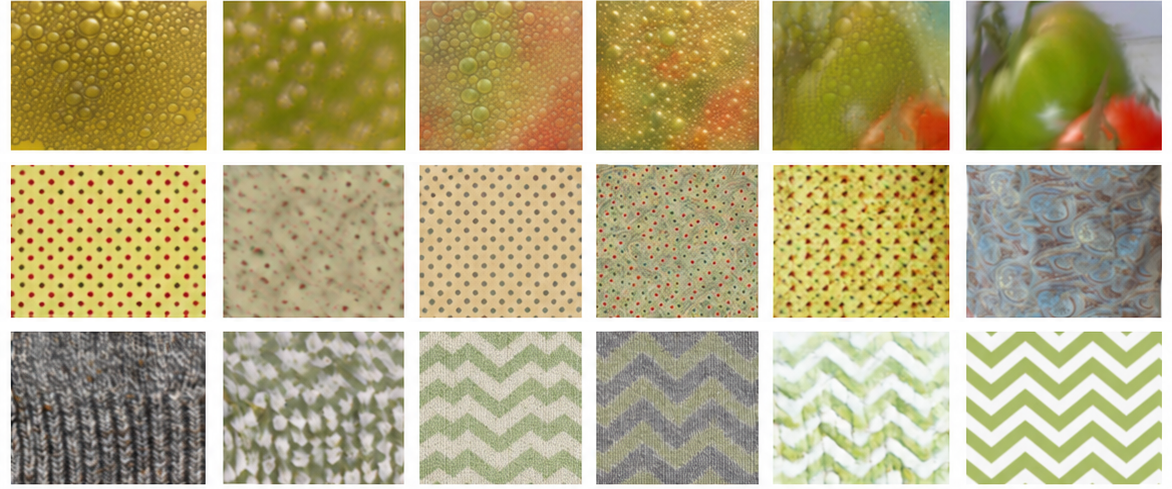}
\caption{The first three pairs of the 120-pair comparison, in pair-ID
order. Columns: $A$, Soft OT (ours), GPT directed, GPT symmetric, Texture Mixer,
$B$. The prompted outputs treat the source arrangement differently from
pair to pair; all eight pairs are in \cref{fig:app-gpt}.}
\label{fig:gptstrip}
\end{figure}

The directed prompt achieves KID \GKIDgpt{}, compared with \GKIDtm{} for
Texture Mixer and \GKIDot{} for soft texton matching
(\cref{tab:generative}); on these 120 pairs soft texton matching lowers
KID against field-linear by \GdKIDotVsPix{} (CI \GdKIDotVsPixCI). Its DINO
realism is close to Texture Mixer's; the symmetric prompt is higher.
Per midpoint, texton matching costs \OTsec{}\,s on one A100 against
\VacherWSec{}\,s for optimization-based interpolation and about
\GPTsec{}\,s of API time for prompted generation; each further $t$ is one
decoder pass (\DecSec{}\,s).
Prompted generation is a strong option for a single plausible intermediate,
and the prompts treat source arrangement differently from pair to pair
(\cref{fig:gptstrip}; all eight pairs in \cref{fig:app-gpt}).

Soft texton matching reuses one fixed source-indexed map, so its path is affine in the
conditioning field; prompted generation prompts each frame independently,
without a shared explicit map.

\subsection{What is paid once, and what is paid per frame?}
\label{sec:runtime}
Encoding and matching are paid once per pair and each coefficient costs one
renderer pass (\cref{tab:runtime}, \cref{app:runtime}). Texture Mixer's
generator is faster than CNT's decoder at every $t$ (\TMInterpSec{}\,s
against \DecSecLow{}\,s at 128\,px); texton matching requires no interpolation-specific training,
training domain, or fixed output size.

%% file: sections/applications.tex
\subsection{Applications}
\label{sec:applications}
\paragraph{Anchor-based texture mixtures.}
Transport maps from one anchor texture to several others give fields on the
anchor's source grid, which can be blended with nonnegative weights summing
to one and decoded
under the anchor's conditioning (\cref{fig:app-mixtures}). The construction
depends on the chosen anchor and is not an exact multi-distribution
barycenter.

\par\medskip\noindent\begin{minipage}{\linewidth}
\centering
\input{figures/mixture_labels}
\includegraphics[width=\linewidth]{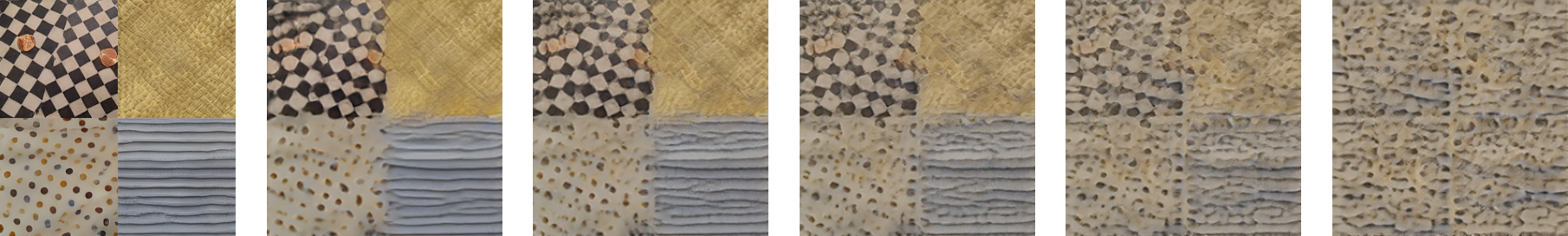}
\captionsetup{hypcap=false}\captionof{figure}{Four-texture mixing with a
common anchor. The input collage identifies the textures; mixing strength
increases from left to right; RT is the re-encoding round-trip error.
Formulation in \cref{app:mixtures}.}
\label{fig:app-mixtures}
\end{minipage}\medskip

\paragraph{Source-conditioned appearance transfer.}
Setting $t=1$ in \cref{eq:path} decodes the transported target appearance
under the source conditioning $L_A$; reversing the two inputs exchanges
those roles, so each texture can be rendered with the other's appearance
(\cref{fig:transfer}).

\par\medskip\noindent\begin{minipage}{\linewidth}
\centering
\begin{minipage}{.72\linewidth}
\input{figures/material_labels}
\includegraphics[width=\linewidth]{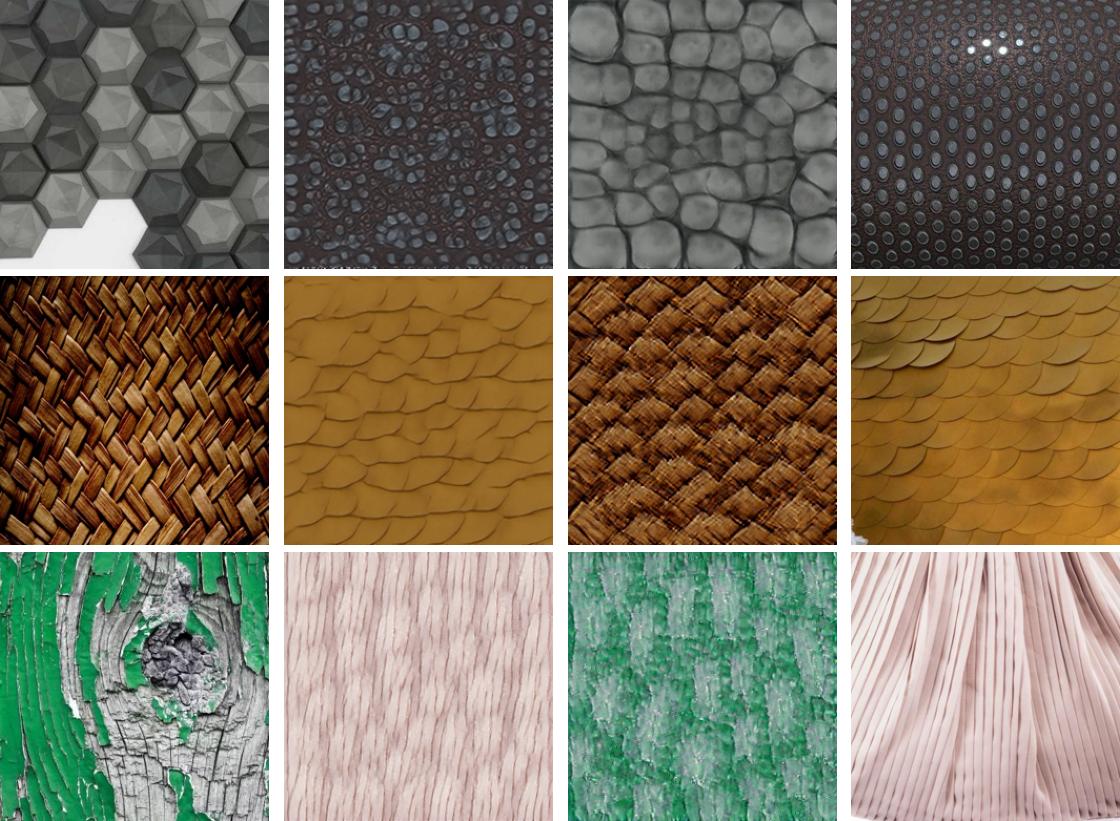}
\end{minipage}
\captionsetup{hypcap=false}\captionof{figure}{Source-conditioned appearance
transfer in both directions, three pairs. Columns: $A$, $B$'s appearance
under $A$'s conditioning, $A$'s appearance under $B$'s conditioning, $B$.
Each output keeps the conditioning texture's arrangement; the operation
does not impose pixelwise structure preservation.}
\label{fig:transfer}
\end{minipage}\medskip

%% file: figures/mixture_labels.tex
\sixpanelheaders{Input collage}{\shortstack{$t=0$\\RT $.14$}}{\shortstack{$t=.25$\\RT $.12$}}{\shortstack{$t=.5$\\RT $.12$}}{\shortstack{$t=.75$\\RT $.16$}}{\shortstack{$t=1$\\RT $.14$}}

%% file: figures/material_labels.tex
\fourpanelheaders{$A$}{\shortstack{$B$ appearance\\$L_A$ conditioning}}{\shortstack{$A$ appearance\\$L_B$ conditioning}}{$B$}

%% file: sections/discussion.tex
\section{Discussion and Limitations}
The task is a directed
transition under retained source conditioning: a diffuse barycentric map
does not reproduce the target distribution exactly, retaining $L_A$ does
not guarantee pixelwise preservation of source structure, and a texture
interpolated with itself is not returned unchanged. Periodic textures remain difficult because
local marginals omit phase and long-range organization, and CNT
reconstruction error bounds the outputs (\cref{app:ceiling}). No single
metric establishes that both endpoints are perceptually represented, and
endpoint balance rewards superposition as readily as realism rewards
typicality; the generative comparator uses single outputs
from fixed prompts; no human preference study is included.

\section{Conclusion}
Texture interpolation is a matching problem before it is a blending problem.
Deciding which textons combine, inside a fixed encoder and decoder, improves
midpoints in CNT and in a trained interpolator; soft matching favours
ensemble statistics, one-to-one matching favours balance.

%% file: sections/statements.tex
\section*{AI Use Statement}
Generative AI tools assisted with manuscript organization and drafting,
literature lookup, analysis of experimental results, mathematical
exposition, figure layout, and LaTeX build and validation scripts, and
assisted with research ideation and experiment execution, including
evaluation code.
GPT-Image-1.5 outputs are evaluated explicitly as a generative baseline.
The experimental results and image assets are our own; editorial assistance
did not generate replacement measurements. Every
number in the main-text tables was recomputed from the saved outputs with
the released scoring code. The authors are responsible for the submitted content.

\section*{Reproducibility Statement}
The appendix specifies the transport objective, numerical settings, Gaussian
map, projection identities, and evaluation protocols. The accompanying
package includes per-pair scores, seeds, solver diagnostics, and pair
manifests for the Texture Mixer matching ablation, the equal-sample ablation,
the alignment study, the timing study, and the transport-cost comparison,
together with the reference image list. The older evaluations retain
aggregate values only. The prompts and the call ledger are in the
repository. Code, the cached features needed to recompute the main-text
tables and the appendix results listed in the repository's check report,
installation scripts, and two Colab notebooks that reproduce the
pipeline end to end are available at
\url{https://github.com/Scientific-Computing-Lab/TextureMatcher}.

\section*{Ethics Statement}
The study concerns texture interpolation and reports no human-subject
experiment. The generated-image baseline and its role in the evaluation are
identified throughout. Texture editing can alter material appearance in ways
that viewers may mistake for captured imagery; edited outputs should be
identified as synthetic when that distinction matters.

%% file: appendix/supplement.tex
\raggedbottom
\section{Protocol and Baseline Definitions}
\label{app:protocol}
The primary benchmark uses 1,000 stratified cross-category pairs from DTD,
which contains 5,640 images in 47 categories~\citep{cimpoi2014dtd}.
Endpoint images are stored at $128\times128$ (Texture Mixer's native
resolution) and bicubically upsampled to $256\times256$ before CNT encoding;
all outputs are resized to $128$\,px before feature extraction. This holds
for the primary, 200-pair matching, and 120-pair generative comparisons.
The reference pool contains
\ReconReference{} DTD images. FID and KID compare midpoint sets with that
pool~\citep{heusel2017fid,binkowski2018kid}; their absolute values change with
the sample count and reference protocol. The older evaluations below retain
their own sample counts and reference definitions.

\paragraph{Two positional baselines.}
Field-linear blends co-located dense CNT feature vectors; our experiment
code names it ``pixel-linear.'' Primitive-linear blends the indexed CNT
elements and is named ``texton-linear'' there. The latter is distinct from CNT's published
interpolation application, which first matches Gaussians by the sum of
differences between centers and weights, then interpolates the matched
parameters~\citep{tu2024cnt}, Appendix B.5. An RGB cross-fade blends image
pixels directly. It appears as a separate reference in the teaser and
parametric-texture experiment. These are three distinct operations.

\paragraph{Transport configuration.}
The finest conditioning field has $256^2$ cells of dimension \FeatureDim.
The default uses $M=\NumSamples$ uniform samples per field, raw feature
vectors, uniform masses, squared Euclidean cost, $\varepsilon=
\SinkhornEpsilon$, and \SinkhornSteps{} Sinkhorn iterations in float32 on
the GPU. The sampled barycentric map is extended through nearest sampled
source vectors. The hard implementation uses 1,024 samples. These different
sample counts are part of the measured implementations. Gaussian transport
uses covariance ridge $\GaussianRidge I$. Our runs draw the samples at
random, and we report the resulting repeat variation below; fixing a sampled
map makes its subsequent path deterministic.

\paragraph{Learned and optimization-based references.}
Texture Mixer uses the released earth-texture checkpoint and its fixed
$128\times128$ generator~\citep{yu2019texturemixer}. DTD is outside that
specific training domain. The Vacher variants optimize Gram, Gaussian, or
Wasserstein statistics~\citep{vacher2020texture}. We run them for 30
iterations; the published configuration uses 150. GPT-Image-1.5 is a
proprietary pretrained generator, evaluated with the image-editing interface
and two reference inputs. No interpolation-specific training is performed
for the CNT operator, which itself relies on pretrained CNT weights.
Timing was re-measured on the 120 pairs under one protocol for all methods
(encode logged once per pair, one renderer pass per $t$); earlier drafts used
a 20-pair run.

\paragraph{KID estimators.}
The CNT block of the 1,000-pair table and the matching-rules table report
KID as the mean over 20 reference subsets; every other KID, and every
$\Delta$KID, uses the full reference with the trace-corrected self-term.

\paragraph{What the metrics condition on.}
Manifold realism is the per-image score of
\citet{kynkaanniemi2019precision}, evaluated in Inception or DINOv2
features~\citep{oquab2024dinov2}. It measures proximity to reference-image
neighborhoods without conditioning on the selected texture pair. We report
medians over five reference-manifold draws; the empirical realism resolution
of the observer experiments is .036, which is not a confidence interval for
FID, KID, or fidelity. DINO balance is the absolute displacement of the endpoint-distance
ratio from one half. LPIPS path measurements use perceptual image
distance~\citep{zhang2018lpips}. Portilla--Simoncelli and BEiTv2 serve as
additional source-fidelity observers~\citep{portilla2000parametric,peng2022beitv2}.

\paragraph{Available evidence.}
The accompanying package includes per-pair scores, seeds, solver diagnostics,
and pair manifests for the Texture Mixer matching ablation, the equal-sample
ablation, the alignment study, the timing study, and the transport-cost
comparison, together with the reference image list. The older evaluations
retain aggregate values only, with confidence intervals where we computed
them. The prompts and the call ledger are in the repository.
The full scoring code is available at the same address.

\ifstandalonesupp\else\clearpage\fi
\section{Operator Derivations}
\label{app:projection}
\subsection{Joint transport and deterministic projection}
For equal-weight sampled clouds, a balanced coupling $P$ defines random
variables $(X,Y)$ with the prescribed empirical marginals. A row-normalized
plan defines $T(X)=\mathbb E[Y\mid X]$. With
$\epsilon_Y=Y-\mathbb E[Y\mid X]$, we have
$\mathbb E[\epsilon_Y\mid X]=0$. Therefore $T(X)$ and $\epsilon_Y$ have zero
cross-covariance, giving
\begin{equation}
\Cov(Y)=\Cov(T(X))+\mathbb E[\Cov(Y\mid X)].
\label{eq:app-covariance}
\end{equation}
The second term is positive semidefinite, so
$\Cov(T(X))\preceq\Cov(Y)$. Equality holds when the conditional variance
vanishes in every direction.

The joint displacement variable $Z_t=(1-t)X+tY$ satisfies
\begin{equation}
\mathbb E[Z_t\mid X]=(1-t)X+tT(X)=\bar Z_t,
\qquad \Cov(Z_t\mid X)=t^2\Cov(Y\mid X).
\end{equation}
Applying total covariance again gives
\begin{equation}
\Cov(Z_t)-\Cov(\bar Z_t)=t^2\mathbb E[\Cov(Y\mid X)].
\end{equation}
This identity describes the cost of replacing the conditional distribution
by its mean before decoding. It imposes no corresponding equality on decoded
image statistics, since the decoder is nonlinear and uses additional
conditioning.

The joint displacement measure
$((1-t)x+ty)_\#P$ reaches both empirical marginals. A diffuse plan can give it
up to $M^2$ support points, whereas the decoder input requires one vector per
source grid cell. Barycentric projection supplies that field but changes the
target distribution. With an unregularized quadratic-cost optimal plan, joint
displacement gives a $W_2$ geodesic~\citep{mccann1997convexity}. Entropic
regularization, projection, nearest-neighbor extension, and nonlinear decoding
each limit the direct use of this geometric interpretation for images.

\subsection{What quadratic cost rewards}
For fixed marginals with means $m_A,m_B$,
\begin{equation}
\mathbb E\|X-Y\|^2=\|m_A-m_B\|^2+
\mathbb E\|X-m_A\|^2+\mathbb E\|Y-m_B\|^2
-2\mathbb E\langle X-m_A,Y-m_B\rangle.
\end{equation}
Unregularized quadratic transport maximizes the cross term. Entropic
transport trades that alignment against plan entropy. A random permutation
preserves the marginals but does not optimize their feature alignment. Coverage and appearance information are therefore separate controls.
Perceptual quality need not follow feature alignment.

\subsection{Gaussian transport}
\label{app:gaussian}
For fitted Gaussian models $\mathcal N(m_A,\Sigma_A)$ and
$\mathcal N(m_B,\Sigma_B)$ with positive-definite covariances, the
quadratic-cost optimal affine map is
\begin{equation}
\begin{aligned}
T_G(x)&=m_B+A(x-m_A),\\
A&=\Sigma_A^{-1/2}
(\Sigma_A^{1/2}\Sigma_B\Sigma_A^{1/2})^{1/2}\Sigma_A^{-1/2}.
\end{aligned}
\end{equation}
The implementation adds $\GaussianRidge I$ to the covariances and clips
negative eigenvalues from floating-point error. CNT field vectors are
mixtures of a much smaller number of primitive features, so covariance
regularization can affect this map. The affine path is the $W_2$ geodesic
for the regularized Gaussian models. The empirical input cloud need not be
Gaussian, and the decoded path retains source conditioning. The Gaussian row is a moment-alignment baseline for the empirical operator.

\subsection{Exact OT (one-to-one)}
A minimum-cost assignment on \CNExact{} equally weighted source and target vectors
is a permutation. Its conditional variance is zero on that sampled cloud.
Nearest-neighbor extension to the dense field can repeat assigned targets,
so sample-level exactness does not imply exact full-field target recovery.
The exact-OT target gap is \HardEndpointGap{} on the 1,000-pair canonical
benchmark; the 200-pair matching-rules study reports \RuleGapHard{} for
exact OT and \RuleGapTransport{} for the default soft plan. A symmetric (OT-sym)
variant averaging the two directed paths reduced order dependence but
lowered recall and set-level statistics, and is not used.

\section{Generative Comparison}
\label{app:generative}
The 120-pair comparison uses a category-stratified subset and the same
$128^2$ scoring resolution for every method. GPT-Image-1.5 receives both
endpoint images in a fixed order and produces one $1024^2$ output per prompt
and pair. The directed prompt requests source layout and halfway target
appearance; the symmetric prompt requests a natural midpoint with no source
preference. Both prompts are reproduced below, byte for byte.

\begin{quote}\small\noindent\textbf{GPT symmetric} (prompt P1):\par
You are given exactly two reference images in order: reference image 1 is texture A and reference image 2 is texture B. Generate one single square image M representing a coherent, plausible perceptual texture midpoint between A and B at t = 0.50.

M must be one internally consistent texture that could plausibly exist on its own. Interpolate texture-level properties such as periodicity, orientation, topology, motif or element shape, spatial scale, density, regularity, material appearance, local contrast, and color distribution. Preserve meaningful evidence of both endpoints and aim for approximately equal perceptual texture distance from A and B. Prefer structural and statistical texture cues over semantic object identity.

Do not make a pixel-space blend, alpha blend, cross-fade, transparent overlay, split image, side-by-side composition, collage, or simple average of colors. Do not merely paste motifs from A and B together. Do not include A, B, labels, text, borders, margins, or a presentation panel. Output only M, edge-to-edge. The definition is symmetric: swapping the order of A and B should not change the intended midpoint.
\end{quote}

\begin{quote}\small\noindent\textbf{GPT directed} (prompt P2):\par
You are given exactly two reference images in order: reference image 1 is source texture A and reference image 2 is target-appearance texture B. Generate one single square midpoint texture M at t = 0.50 for a directed source-layout appearance interpolation from A toward B.

Preserve the dominant spatial organization, layout, orientation, and large-scale arrangement of A, while moving the appearance of its local texture elements halfway toward B. Interpolate local motif shape, scale, density, regularity, material, local contrast, and color statistics toward B. The result must be one coherent, plausible texture, not two overlaid patterns. Semantic objects that are incidental to texture should not dominate the interpolation.

Do not make a pixel-space blend, alpha blend, cross-fade, transparent overlay, split image, side-by-side composition, collage, or simple average of colors. Do not include A, B, labels, text, borders, margins, or a presentation panel. Output only M, edge-to-edge. This task is directed: A supplies layout and B supplies the target appearance.
\end{quote}

Model gpt-image-1.5 through the image-editing interface, quality high,
input fidelity high, two ordered reference images ($A$ then $B$), one
$1024\times1024$ output per call; 240 scored calls, median latency
\GPTsec{}\,s.

\begin{table}[ht]
\centering\small
\caption{Complete 120-pair results. FID is descriptive at this
sample count. Bold marks the best value in each column.}
\label{tab:app-generative}
\setlength{\tabcolsep}{4.8pt}
\begin{tabular}{lccccc}
\toprule
Method & FID $\downarrow$ & KID $\downarrow$ & Inc. real. $\uparrow$ & DINO real. $\uparrow$ & Balance $\downarrow$ \\
\midrule
Field-linear & \GFIDpix & \GKIDpix & \GRIncpix & \GRDinpix & \GPlacepix \\
Primitive-linear & \GFIDtex & \GKIDtex & \GRInctex & \GRDintex & \textbf{\GPlacetex} \\
Gaussian $W_2$ & \GFIDgauss & \GKIDgauss & \GRIncgauss & \GRDingauss & \GPlacegauss \\
Soft OT (ours) & \GFIDot & \GKIDot & \GRIncot & \GRDinot & \GPlaceot \\
Exact OT & \GFIDhard & \GKIDhard & \GRInchard & \GRDinhard & \GPlacehard \\
Texture Mixer & \GFIDtm & \GKIDtm & \GRInctm & \GRDintm & \GPlacetm \\
\midrule
GPT, directed & \textbf{\GFIDgpt} & \textbf{\GKIDgpt} & \textbf{\GRIncgpt} & \GRDingpt & \GPlacegpt \\
GPT, symmetric & \GFIDgptsym & \GKIDgptsym & \GRIncgptsym & \textbf{\GRDingptsym} & \GPlacegptsym \\
\bottomrule
\end{tabular}
\end{table}

The directed prompt's median distance-ratio placement is \GPlaceMedgpt.
This is a signed location along the endpoint-distance axis, distinct from
the absolute balance error in the table. Paired DINO realism differences
from the CNT-based variants range from \GgptDRDinMin{} to \GgptDRDinMax{}
($p\le10^{-5}$); the difference from Texture Mixer is \GgptDRDinTM{} and not
significant. Paired differences and differences between separately
summarized table columns need not coincide. Soft texton matching reuses a fixed
source-indexed map and has explicit affine dependence on $t$ before
decoding, while independently prompted frames carry no such structure; its
dependence on input order remains. The API latency is approximately
\GPTsec\,s per output, versus \OTsec\,s for a local soft texton matching midpoint in the
A100 timing study.

\begin{figure}[p]
\centering
\sixpanelheaders{$A$}{Soft OT (ours)}{GPT directed}{GPT symmetric}{Texture Mixer}{$B$}
\includegraphics[width=\linewidth]{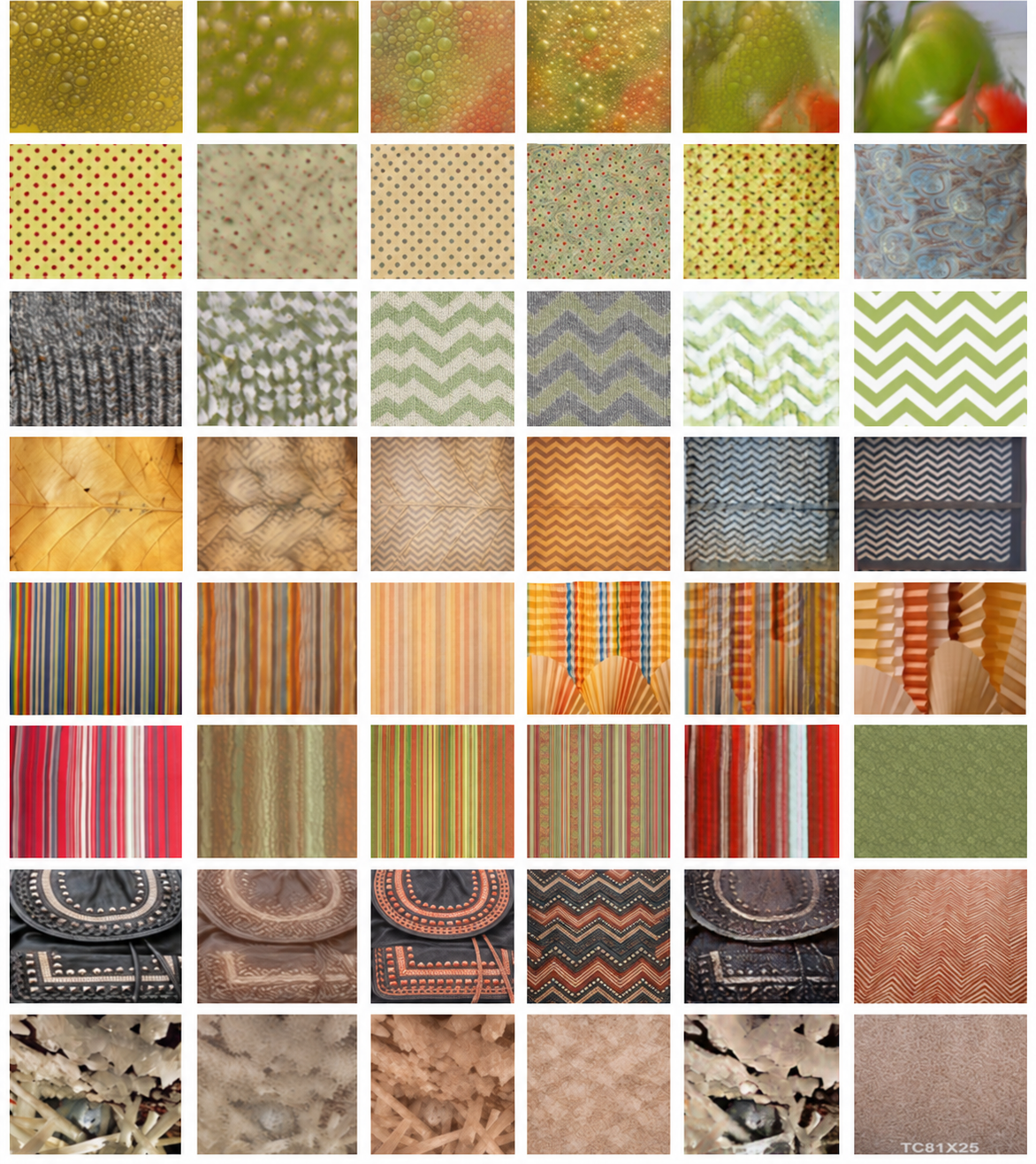}
\caption{The first eight pairs, in pair-ID order.
Columns are $A$, Soft OT (ours), GPT directed, GPT symmetric, Texture Mixer,
and $B$; the main text shows the first three rows. The directed prompt handles stochastic appearance, structured motifs, and
semantic source content differently; the examples do not establish a
pixelwise structure guarantee for either method.}
\label{fig:app-gpt}
\end{figure}
\FloatBarrier

\section{Additional Quantitative Evaluations}
\subsection{Optimization-based comparison}
\begin{table}[ht]
\centering\small
\caption{The 50-pair comparison with optimization-based interpolation,
scored at 128\,px. Timing is on one A100, with 30 iterations for Vacher
variants. FID at this sample count is descriptive. --: not defined, no
interpolation is performed.}
\label{tab:app-vacher}
\begin{tabular}{lccc}
\toprule
Method & FID $\downarrow$ & Midpoint (s) & 11-frame path (s) \\
\midrule
Soft OT (ours) & 268.3 & \OTsec & \PathSecOT \\
Field-linear & 284.6 & \LinearSec & \PathSecLinear \\
Vacher, Gram & 318.4 & 13.5 & 148.5 \\
Vacher, Gaussian & 314.3 & 15.0 & 165.3 \\
Vacher, Wasserstein & 313.6 & \VacherWSec & \TimeVacherPath \\
CNT reconstruction & 273.9 & -- & -- \\
\bottomrule
\end{tabular}
\end{table}
We find no significant paired difference in per-sample realism and midpoint
balance (all Wilcoxon $p>.25$). This small-sample result does not establish
statistical equivalence. The runtime difference follows from reusing the
transport map across frames, while the optimization-based method solves for
each output independently. The Wasserstein implementation takes 52.1, 52.4,
and 56.5 seconds at 128, 256, and 512 pixels, respectively.

\FloatBarrier
\subsection{Matching on CNT's Gaussian primitives}
\label{app:primitives}
CNT predicts about \NumGaussians{} Gaussian primitives per texture, of
which about \GEPrimN{} carry most of the weight. On 120 pairs we pair the two primitive sets with a Hungarian
assignment under a position cost, an appearance cost, or their mixture, and
either freeze the source layout or let the matched primitives move; the
encoder, primitives, interpolation, and decoder are otherwise those of the
main protocol (\cref{tab:app-primitives}). With a frozen layout the cost is
a null result (KID \GEKIDfixP{} / \GEKIDfixA{} / \GEKIDfixPA{} for position /
appearance / mixed; appearance against position \GEdKIDfix{}, CI
\GEdKIDfixCI), and reassigning every pairing changes per-image realism by
\GEScrambleDReal. With a moving layout, appearance matching is the worst
choice (\GEKIDmorphA{} against \GEKIDmorphP, $\Delta$KID \GEdKIDmorph{}, CI
\GEdKIDmorphCI), because look-alike partners are far apart and the moved
primitives superpose. Intervals are paired leave-one-pair-out jackknife,
95\% CI. A soft matching of the same primitives reaches
\GEKIDprimSoft{} KID, against \GEKIDdenseOT{} for the dense operator and
\GEKIDprimHard{} for every one-to-one rule. The matching's softness and the
anchoring to the source layout, not the cue used to pair, separate a coherent
intermediate from a superposition.

\begin{table}[ht]
\centering\small
\caption{KID for one-to-one matchings of CNT's Gaussian primitives, 120
pairs, canonical scoring. Rows fix or move the source layout; columns give
the pairing cost.}
\label{tab:app-primitives}
\begin{tabular}{lccc}
\toprule
Layout & Position cost & Appearance cost & Mixed cost \\
\midrule
Frozen (source) & \GEKIDfixP & \GEKIDfixA & \GEKIDfixPA \\
Moving (matched) & \GEKIDmorphP & \GEKIDmorphA & \GEKIDmorphPA \\
\bottomrule
\end{tabular}
\end{table}

\subsection{Matching rules under the original rendering protocol}
\label{app:rules}
The 200-pair matching-rule study of the main text, scored under the
original rendering protocol, is tabulated here.

\begin{table}[ht]
\centering\small
\caption{Matching rules on the same 200 pairs. Random permutation preserves
the sampled marginals but ignores appearance. Mean and nearest-neighbor maps
test averaging and target reuse. Lower KID/FID is better; gaps are normalized
target-end distances (the positional rows reach the target field by
construction, so their gap is decoder error only). Bold marks the best value
per column; this table uses the original rendering protocol,
unlike the matched-rendering table in the main text.
Bottom line: every soft appearance-aware rule beats positional pairing on
KID, and entropic transport outperforms exact OT (one-to-one) in the
matched-rendering comparison (\ifstandalonesupp main text\else\cref{tab:eqsample}\fi),
so the form of the matching matters, not appearance information alone.}
\label{tab:rules}
\setlength{\tabcolsep}{6pt}
\begin{tabular}{lccccc}
\toprule
Rule & KID $\downarrow$ & FID $\downarrow$ & $b$ $\downarrow$ & Target gap $\downarrow$ & Inc. real. $\uparrow$ \\
\midrule
Random permutation & \RuleKIDRandom & \RuleFIDRandom & \textbf{\RuleBalRandom} & \RuleGapRandom & \textbf{\RandomIncRealism} \\
Target mean & \RuleKIDMean & \RuleFIDMean & \RuleBalMean & \RuleGapMean & \RuleIncMean \\
Field-linear & \RuleKIDPixel & \RuleFIDPixel & \RuleBalPixel & \RuleGapPixel & \RuleIncPixel \\
Primitive-linear & \RuleKIDPosition & \RuleFIDPosition & \textbf{\RuleBalPosition} & \textbf{\RuleGapPosition} & \RuleIncPosition \\
Gaussian $W_2$ & \RuleKIDGaussian & \RuleFIDGaussian & \RuleBalGaussian & \RuleGapGaussian & \RuleIncGaussian \\
Nearest neighbor & \RuleKIDNN & \RuleFIDNN & \RuleBalNN & \RuleGapNN & \RuleIncNN \\
Exact OT & \RuleKIDHard & \RuleFIDHard & \RuleBalHard & \RuleGapHard & \RuleIncHard \\
Soft OT (ours), $\varepsilon=.05$ & \RuleKIDTransport & \RuleFIDTransport & \RuleBalTransport & \RuleGapTransport & \RuleIncTransport \\
Soft OT (ours), $\varepsilon=.005$ & \textbf{\RuleKIDSharp} & \textbf{\RuleFIDSharp} & \RuleBalSharp & \RuleGapSharp & \RuleIncSharp \\
\bottomrule
\end{tabular}
\end{table}

\subsection{Equal-sample comparison of hard and soft matchings}
\label{app:equal}
The main-text hard and entropic rows use \CNExact{} and \SubsampleM{}
samples. The matched-rendering ablation
(\ifstandalonesupp main-text table\else\cref{tab:eqsample}\fi) repeats the
200-pair study under one rendering path. Exact OT (one-to-one) and the two \CNExact{}-sample entropic rows
share the same sampled clouds; field-linear uses positional correspondence
on the full co-located fields; the \SubsampleM{}-sample operator uses its
own sample; all rows share the field-blend and pyramid construction and were
scored in one session.
Sinkhorn is iterated until the marginal error is below $\SinkhornMargTol$.
The sharp plan retains \EqVarRetSharp{} of the target variance and the
default plan \EqVarRet. Against field-linear on the same pipeline (KID
\EqKIDlinear), the default plan gives \EqdKIDotVsLinear{} (CI
\EqdKIDotVsLinearCI), the \SubsampleM{}-sample operator
\EqdKIDotPaperVsLinear{} (CI \EqdKIDotPaperVsLinearCI), and exact OT
(one-to-one) \EqdKIDhardVsLinear{} (CI \EqdKIDhardVsLinearCI). The sharp plan's difference from the default plan is
\EqdKIDsharp{} (CI \EqdKIDsharpCI); the difference of exact OT (one-to-one) from
the sharp plan is \EqdKIDhardSharp{} (CI \EqdKIDhardSharpCI) and from the
default plan \EqdKIDhard{} (CI \EqdKIDhardCI). All $\Delta$KID intervals are
paired leave-one-pair-out jackknife, 95\% CI. The balance and target-gap
differences between hard and default are significant at $p=\EqBalP$ and
$p=\EqGapP$ (paired Wilcoxon). In this matched-rendering study the
normalized target-end gap at $t=1$ is \EqGaphard{} for exact OT, \EqGapotSharp{} for the sharp plan and
\EqGapot{} for the default plan; it is not defined for field-linear, whose
endpoint is the target field by construction, and was not measured for the
\SubsampleM{}-sample operator. Re-encoding intermediates at
$t\in\{.25,.5,.75\}$ gives error \RoundTripTransport{} against a
\RoundTripFloor{} reconstruction floor, versus \RoundTripPosition{} for
field-linear blending.

\begin{table}[ht]
\centering\small
\caption{Feature spread retained at the midpoint, $n=40$.
The common direction across representations supports the matching mechanism.
In a separate experiment decoding DINO features to images, appearance-aware
matching improves LPIPS midpoint placement on \DINOPlacementWins{} of the
$n=40$ pairs.}
\label{tab:app-representation}
\begin{tabular}{lcc}
\toprule
Representation & Positional blend & Appearance-aware matching \\
\midrule
DINO & \SpreadDINOLin & \SpreadDINOTransport \\
SAM & \SpreadSAMLin & \SpreadSAMTransport \\
CNT primitives & \SpreadCNTElementLin & \SpreadCNTElementTransport \\
CNT field & \SpreadCNTFieldLin & \SpreadCNTFieldTransport \\
\bottomrule
\end{tabular}
\end{table}

\subsection{Matching in Texture Mixer's local latent}
\label{app:tmcoupling}
Texture Mixer interpolates a $32\times32$ local latent positionally and
blends a global latent. For the matching ablation
(\ifstandalonesupp the main-text table\else\cref{tab:canonical}\fi) we keep its encodings,
global-latent blend, and generator, and replace only the pairing of the two
local-latent clouds. The entropic regularization is set as a multiple of
the median pairwise cost, $\varepsilon=\TMEpsRel{}\times$median, and
Sinkhorn is run until the marginal error is below $\SinkhornMargTol$. At that
setting the projected cloud retains \TMVarRetOt{} of the target variance;
in the diffuse limit the coupling reduces to the target-mean rule; at the
reported setting it retains \TMVarRetOt{} of the target variance, and exact
OT (one-to-one) retains the full sampled target cloud. Scoring is identical to
the main protocol at 128\,px against the same reference pool: the native,
target-mean and soft rows are scored on \TMkN{} pairs, the one-to-one row on
the 120-pair subset. $\Delta$KID against the native positional blend is a
paired leave-one-pair-out jackknife, 95\% CI, for every row; on \TMkN{}
pairs the soft matching differs from the target mean by \TMkdKIDotVsMean{}
(CI \TMkdKIDotVsMeanCI), and its per-image realism exceeds the native
blend's ($p=\TMkRealP$).

\subsection{Encoding ceiling}
\label{app:ceiling}
This study uses a separate reference of 5,640 native 128-px DTD windows.
One CNT encode--decode loop has LPIPS \CeilLPIPSrec{} on the 120-pair set.
The KID of the originals is \CeilKIDorig, measured against this study's
5,640-window reference, so these values are not comparable to
\ifstandalonesupp the main-text 120-pair table's\else\cref{tab:generative}'s\fi; their reconstructions reach \CeilKIDrec, and a
second reconstruction \CeilKIDrecTwo. Soft texton matching midpoints
are at \CeilKIDmid, below the double-reconstruction level. Per-image realism
moves from \CeilRealOrig{} for originals to \CeilRealRec{} for
reconstructions and \CeilRealMid{} for midpoints. Per-pair reconstruction
error does not predict midpoint quality (Spearman \CeilRho).

\subsection{Observer dependence and negative controls}
An older 100-pair realism experiment includes an LPIPS-optimized control.
It attains a near-perfect LPIPS midpoint ratio (.50), despite producing a
degenerate image. Its realism is below that of soft texton matching, though not below every
baseline. The random-pairing control in the main paper has high per-image realism
despite poor set-level statistics, illustrating the converse problem.

\begin{table}[ht]
\centering\small
\caption{Per-image realism, $n=100$. $R\ge1$ denotes inclusion in the
reference manifold estimate. The paired win column compares soft texton matching to
each row; --: not defined for its own row.}
\label{tab:app-realism}
\begin{tabular}{lccc}
\toprule
Method & Mean $R$ & Fraction $R\ge1$ & Soft OT wins ($p$) \\
\midrule
Primitive-linear & .920 & 14\% & 65/100 ($2.6\times10^{-4}$) \\
Field-linear & .952 & 29\% & 54/100 (.21) \\
LPIPS control & .938 & 25\% & 62/100 (.017) \\
Soft OT (ours) & .960 & 41\% & -- \\
\bottomrule
\end{tabular}
\end{table}

In a separate 1,000-pair observer evaluation, Texture Mixer and soft texton matching have
Inception realism .963 and .962 (paired $p=.28$), respectively. Their
Portilla--Simoncelli fidelity scores are $-.575$ and $-.585$ ($p=.18$),
while DINO realism is .948 versus .904 and BEiTv2 fidelity is $-.875$ versus
$-.906$. These measurements belong to the earlier observer protocol and remain
separate from the canonical rerun.

\subsection{Endpoint direction}
The operator is directed: at $t=0$ the feature field is exactly the source
field, so a nonzero image gap there measures the decoder's reconstruction
error, and swapping the two inputs changes the output because the target's
appearance is decoded under the source's conditioning.
\FloatBarrier
\subsection{Regularization and repeated application}
For the 20-pair entropy sweep, mean row entropy changes from
\EntropyLow{} to \EntropyHigh{} nats between $\varepsilon=.005$ and $.5$,
against a uniform ceiling of \EntropyCeiling. Projected target variance
changes from \VarianceLow{} to \VarianceHigh. The extreme-midpoint LPIPS
difference is .21 (IQR .17--.25), compared with an approximately .15
resampling floor. These are midpoint measurements; the path covariance
identity explains why projection can have larger effects toward $t=1$.

\begin{figure}[ht]
\centering
\includegraphics[width=\linewidth]{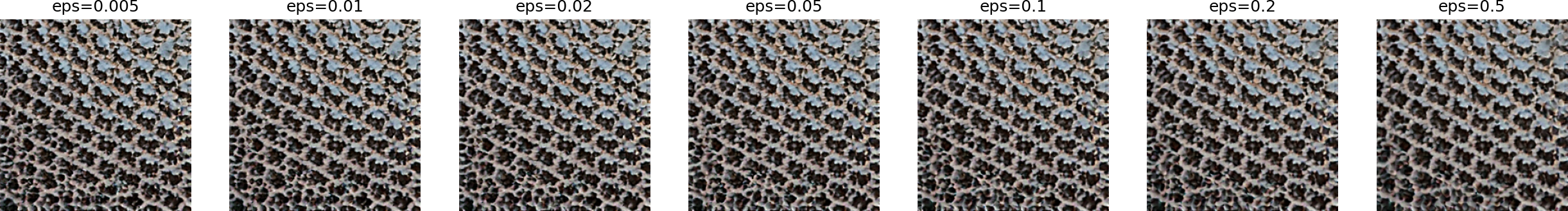}
\caption{Complete regularization sweep for one pair. The
comparatively small visual change at the midpoint accompanies much larger
changes in the entropic plan and projected feature variance.}
\label{fig:app-epsilon}
\end{figure}

\begin{table}[ht]
\centering\small
\caption{Eight repeated operations on 24 pairs. Self: use the current image
on both sides. Fixed: retain a fixed second endpoint. Positive realism drop
denotes deterioration under that observer; negative values denote an increase.
--: not defined, no second endpoint.}
\label{tab:app-repeat}
\begin{tabular}{lccc}
\toprule
Method & Realism drop (self) & Realism drop (fixed) & Self drift \\
\midrule
Autoencoder only & .105 & -- & 61.9 \\
Primitive-linear & .024 & .025 & 61.0 \\
Field-linear & .024 & .029 & 61.0 \\
Texture Mixer & $-.019$ & .006 & 59.7 \\
Soft OT (ours) & $-.041$ & $-.077$ & 62.8 \\
\bottomrule
\end{tabular}
\end{table}
Autoencoder-only and transport self loops have similar DINO drift. Increasing
realism along these trajectories therefore does not imply preservation of
the original image; the outputs can move toward smoother, typical textures.
\FloatBarrier

\subsection{Runtime}
\label{app:runtime}
\Cref{tab:runtime} separates what is paid once per pair from what each
interpolation coefficient costs. The optimization-based comparison pays for
each frame separately, and the GPT entry is API latency rather than a
hardware-normalized runtime.

\begin{table}[ht]
\centering\small
\caption{Recorded stage timings in seconds on one A100: 120 pairs for the
CNT and Texture Mixer rows; the Vacher \etal timing comes from a 20-pair run
on the same hardware with its 30-iteration configuration; GPT is the median
API latency over the scored calls. Encoding is separate; each interpolation
coefficient is one renderer pass. CNT decoding is at 256\,px (128\,px in
parentheses); Texture Mixer is at 128\,px. --: no such stage for this method.}
\label{tab:runtime}
\setlength{\tabcolsep}{6pt}
\begin{tabular}{lccc}
\toprule
Method & Encode, once per pair & Couple, once per pair & Each requested $t$ \\
\midrule
Field-linear & \EncodeSec & -- & \DecSec{} (\DecSecLow) decode \\
Soft OT (ours) & \EncodeSec & \OTTransportSec & \DecSec{} (\DecSecLow) decode \\
Exact OT & \EncodeSec & \HardAssignSec & \DecSec{} (\DecSecLow) decode \\
Texture Mixer & \TMEncSec & -- & \TMInterpSec{} generate \\
Vacher $W_2$ & -- & -- & \VacherWSec{} optimize \\
GPT-Image-1.5 & -- & -- & $\sim\GPTsec$ API call \\
\bottomrule
\end{tabular}
\end{table}
\FloatBarrier

\clearpage
\section{Qualitative Comparisons and Applications}
\label{app:qualitative}
\begin{figure}[ht]
\centering
\sixpanelheaders{$A$}{CNT-linear}{Gaussian $W_2$}{Soft OT (ours)}{Texture Mixer}{$B$}
\includegraphics[width=\linewidth]{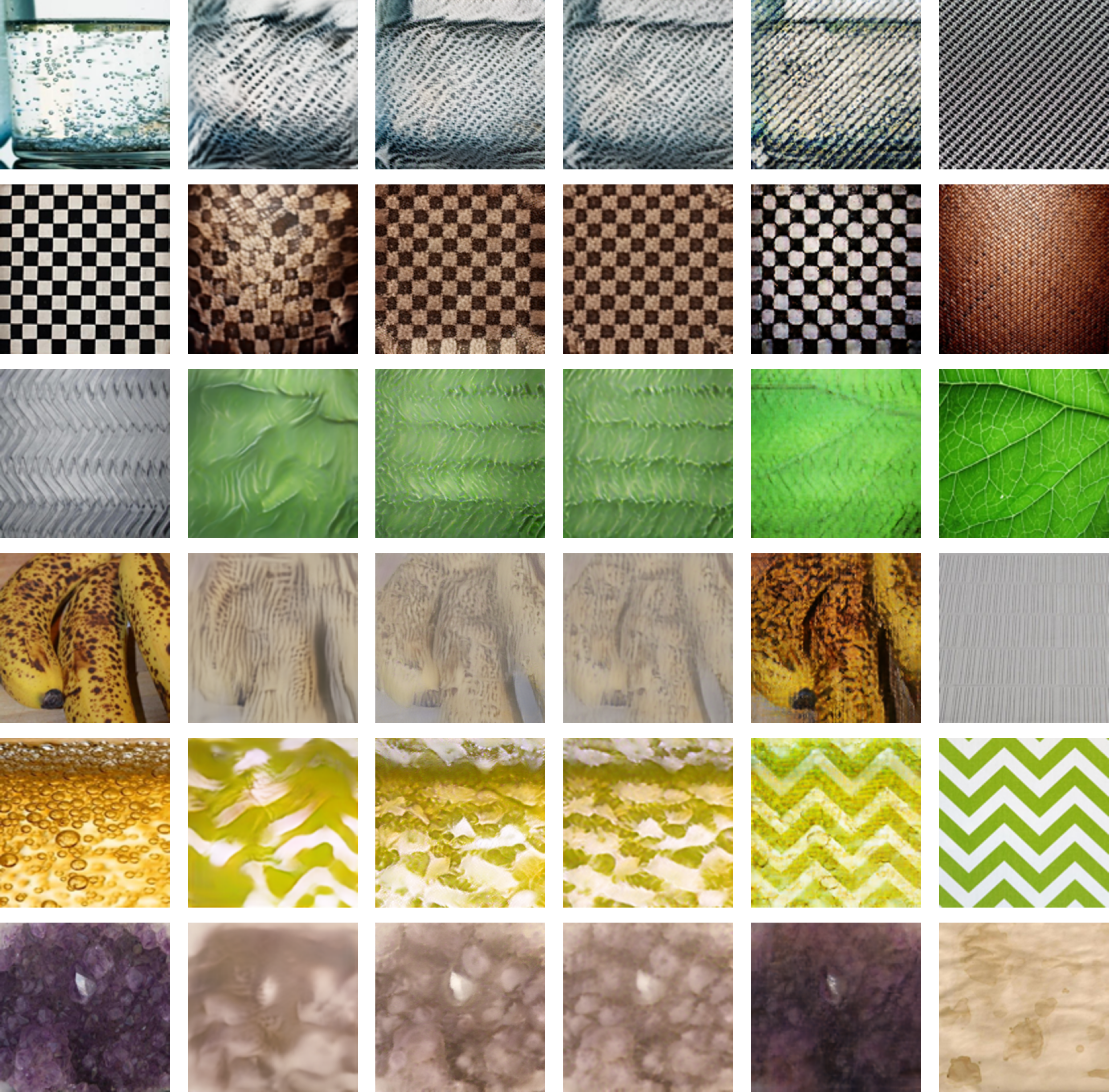}
\caption{The full six-pair qualitative comparison. The main
paper selects rows 2, 4, and 5. The Gaussian and empirical columns often
resemble one another, while the relative appearance of all methods depends
on the pair.}
\label{fig:app-qualitative}
\end{figure}
\FloatBarrier

\begin{figure}[ht]
\centering
\begin{minipage}{.87\linewidth}
\fourpanelheaders{$A$}{Positional linear}{Soft OT (ours)}{$B$}
\includegraphics[width=\linewidth]{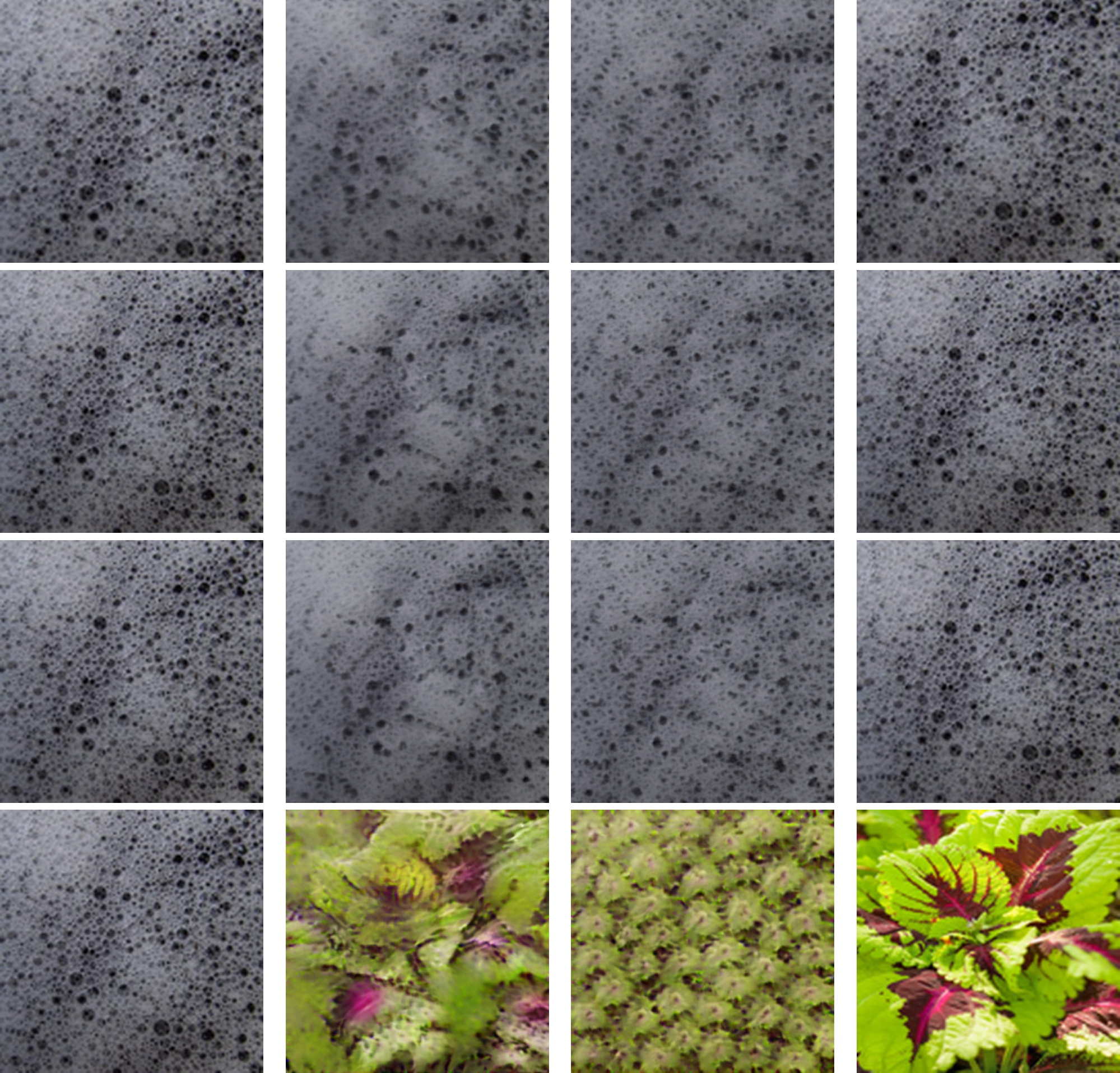}
\end{minipage}
\caption{Full correspondence progression. Columns: $A$, positional linear,
Soft OT (ours), $B$. Rows: overlapping crops, a $2^\circ$ warp, an $8^\circ$
warp, and an unrelated pair. Spatial pairing remains useful in the aligned
rows; appearance matching gives a coherent intermediate in the unrelated pair.}
\label{fig:app-boundary}
\end{figure}
\FloatBarrier

\begin{figure}[ht]
\centering
\includegraphics[width=.8\linewidth]{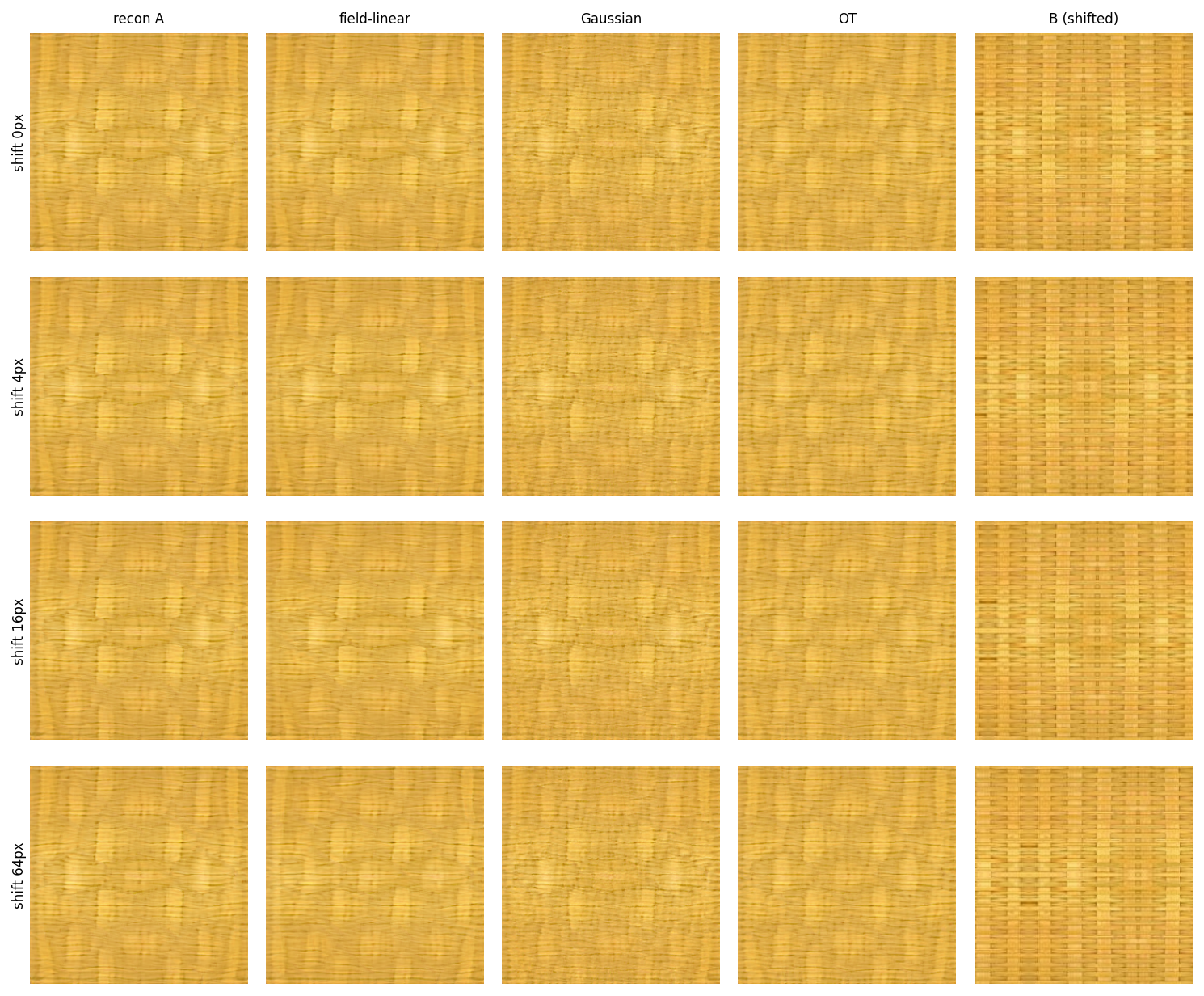}
\caption{Alignment isolated. Rows: cyclic shift of $B$ by 0, 4, 16, 64\,px;
columns: source reconstruction, field-linear, Gaussian, Soft OT (ours), $B$.
Positional blending superposes as the shift grows; the appearance-aware
rules are far less sensitive to displacement.}
\label{fig:app-alignshift}
\end{figure}
\FloatBarrier

\paragraph{Source-conditioned material transfer.}
At $t=1$, the operator uses target-aligned appearance with source conditioning.
Reversing the arguments reverses those roles. The examples in
the main-text transfer figure illustrate the resulting control over
arrangement; the operation does not impose a pixelwise geometry constraint.

\paragraph{Spatially varying paths.}
The interpolation coefficient in a frozen map can vary across source cells:
$z_i=(1-t_i)x_i+t_iT(x_i)$. A scalar ramp produces a spatial transition;
an externally chosen field gives spatial control. This extension changes
the mixing coefficients without requiring another transport solve.
\FloatBarrier

\section{Exploratory Mixtures, Coding, and Plan Diagnostics}
\subsection{Anchor-based mixtures}
\label{app:mixtures}
\begin{figure}[ht]
\centering
\fivepanelheaders{Input collage}{$t=.25$}{$t=.5$}{$t=.75$}{$t=1$}
\includegraphics[width=\linewidth]{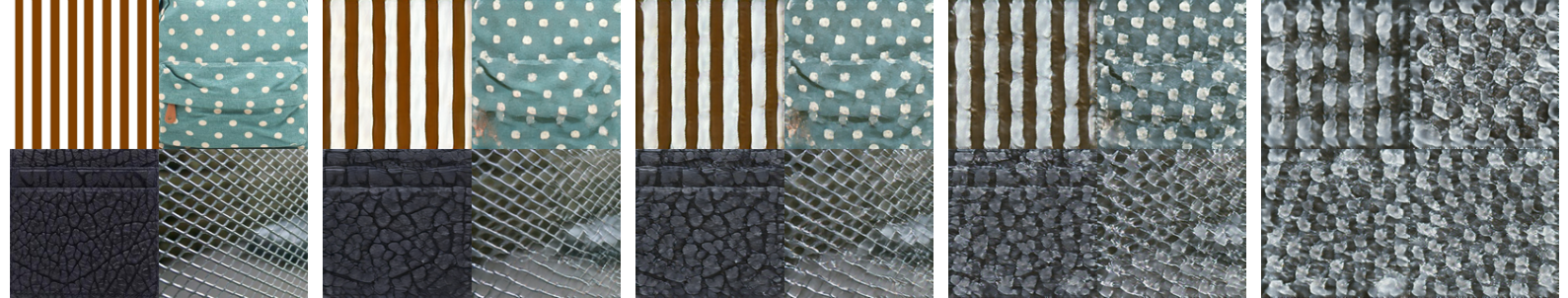}
\caption{A second four-texture example: a $2\times2$ collage of four
textures moved toward their anchor-based mixture as $t$ increases.}
\label{fig:app-mixtures2}
\end{figure}

\begin{figure}[ht]
\centering
\includegraphics[width=.3\linewidth]{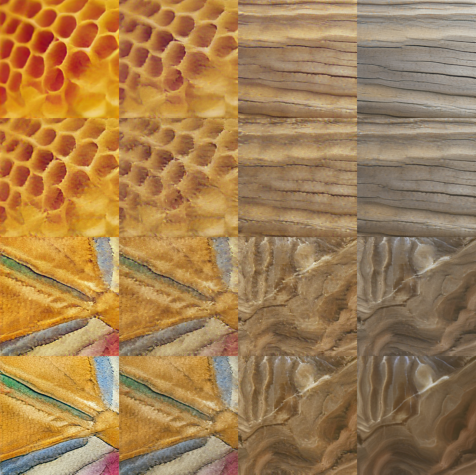}
\caption{A $4\times4$ palette spanned by four corner textures under one
anchor; every cell is a decoded mixture with the same conditioning.}
\label{fig:app-palette}
\end{figure}

\begin{figure}[ht]
\centering
\sixpanelheaders{Regions ($k=3$)}{$t=0$}{$t=.25$}{$t=.5$}{$t=.75$}{$t=1$}
\includegraphics[width=\linewidth]{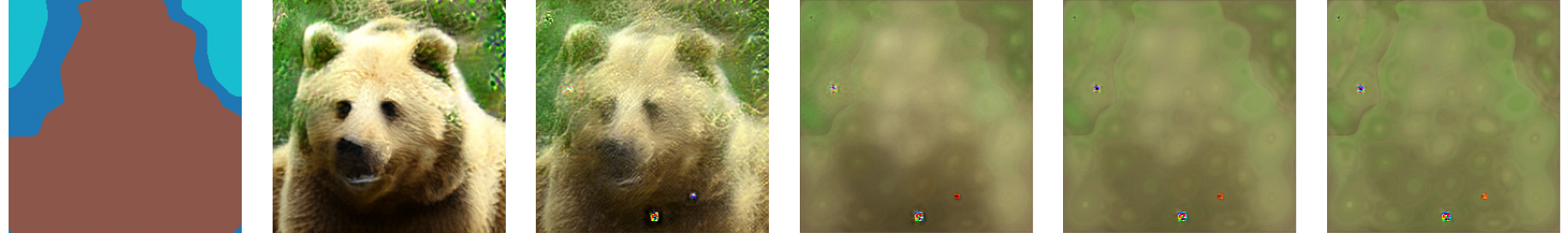}
\caption{Exploratory: a natural image clustered into three regions and
moved toward their anchor-based mixture. Objects dissolve as distributions;
the construction is not an exact barycenter and is outside the texture
setting of the paper.}
\label{fig:app-semantic}
\end{figure}

For $k$ textures, choose source $X_1$ as an anchor and form
\begin{equation}
X_w=\sum_{j=1}^{k}w_jT_{1\to j}(X_1),\qquad
w\in\Delta_k,\quad T_{1\to1}=\mathrm{id}.
\end{equation}
The decoder uses the anchor's conditioning. This one-step construction is
anchor-dependent. On two four-texture sets, changing the anchor changes the
output by \AnchorChangeA{} and \AnchorChangeB{} of mean source separation, versus same-anchor rerun
floors of 14\% and 32\%. Along a four-texture path, round-trip
errors are .12--.16 against a .14 reconstruction floor. These mixtures depend on an anchor and differ from exact multi-distribution
Wasserstein barycenters. The four-texture strip is in the main text
(\ifstandalonesupp Applications subsection\else\cref{fig:app-mixtures}\fi);
its RT labels are the re-encoding round-trip errors quoted above.
\Cref{fig:app-mixtures2} is a second four-texture example,
\cref{fig:app-palette} browses the same construction as a $4\times4$
palette, and \cref{fig:app-semantic} applies it to a natural image split
into three regions, where the objects dissolve into a mixture rather than a
recognizable scene; none is an exact barycenter.

\subsection{Sparse appearance coding}
\begin{figure}[ht]
\centering
\includegraphics[width=.9\linewidth]{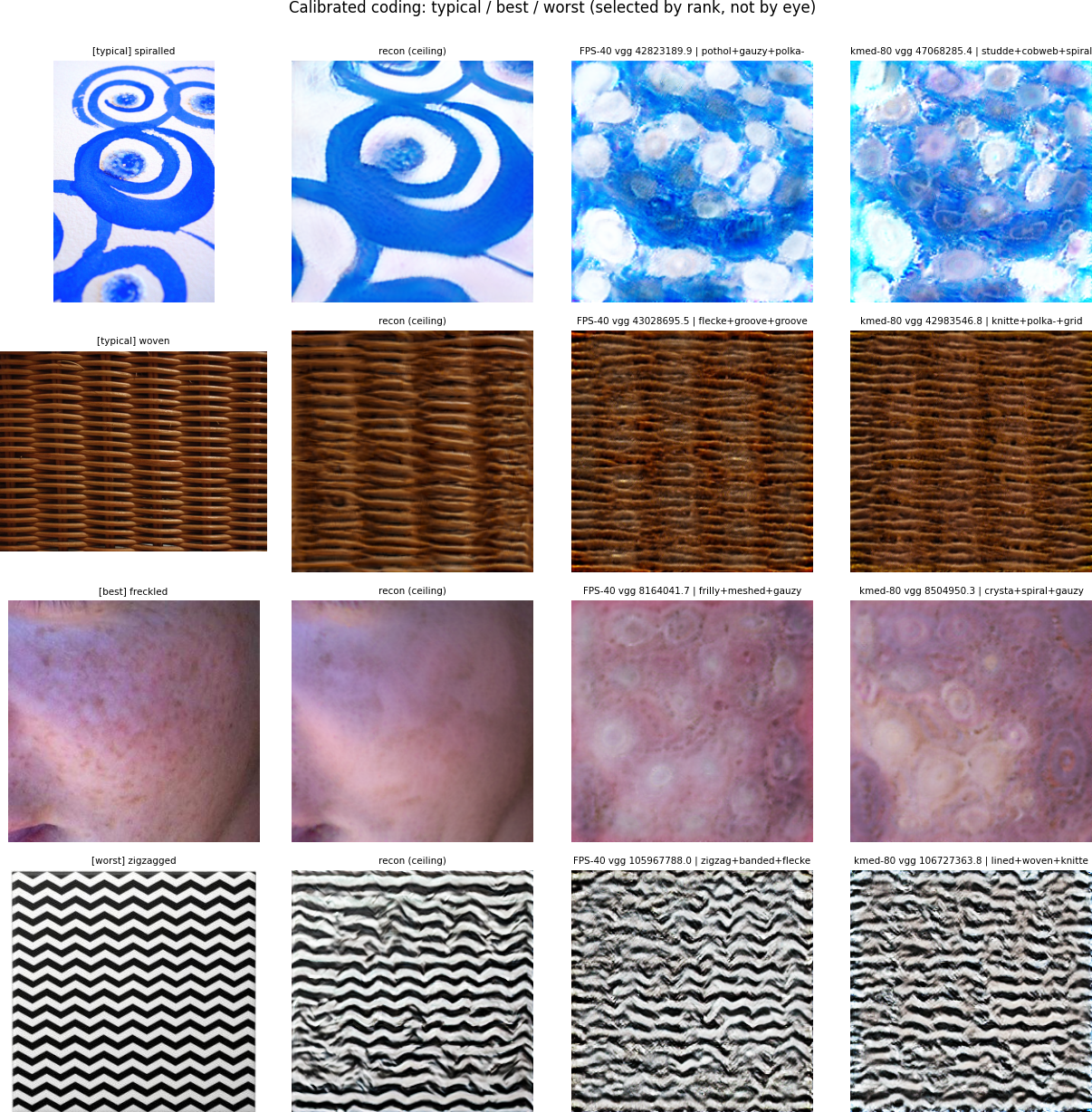}
\caption{Exploratory dictionary reconstructions. Columns show the target,
CNT reconstruction, and decoded constructions from two dictionary choices.
The bottom row illustrates the difficulty of preserving regular structure.}
\label{fig:app-coding}
\end{figure}

For target field $Z$ and dictionary textures $A_j$, fixed transported atoms
$T_{Z\to A_j}(Z)$ share the target's support. The weight problem is
\begin{equation}
\min_{w\in\Delta_K}
\left\|\sum_j w_jT_{Z\to A_j}(Z)-Z\right\|_F^2.
\end{equation}
It is a convex quadratic problem for fixed atoms; uniqueness requires
additional conditions. Sparse subset selection is a separate combinatorial
step. Among 83 reconstructable targets and a 40-atom dictionary, the recorded
sparse mixture improves on the best single atom in 82\% of cases, with a
median margin of 22\%. Sixteen of 99 targets were excluded because CNT could
not reconstruct them. Periodic structure is a recurring failure
(\cref{fig:app-coding}).

The dictionary study uses a pool of 376 textures. Raw farthest-point sampling
can preferentially select corrupted encodings. Screening and a
landmark-completed geometry improve that selection procedure; screened FPS
with 40 atoms matches the decoded quality of 80 $k$-medoids atoms
($p=.42$). This result does not establish superiority of a selection rule.
We did not retain the subset-search settings and raw coding outputs, so
coding remains an exploratory extension.

\subsection{Plan alignment}
With centered clouds and RMS amplitudes $a_A,a_B$, define
\begin{equation}
\rho(P)=\frac{1}{a_Aa_B}\sum_{ij}P_{ij}
\langle x_i-\bar x,y_j-\bar y\rangle.
\end{equation}
For nonzero amplitudes this lies in $[-1,1]$ by Cauchy--Schwarz. A sharp
1,024-point assignment on 1,000 screened pairs yields 929 pairs at
$\RhoHealthy$ and 71 at $\RhoFlagged$. The threshold was selected on the
same data. Sharp entropic and hard-plan scores have correlation
.99; the default diffuse plan is less informative. This describes pair geometry in-sample; prospective failure detection has
not been validated.

\begin{figure}[ht]
\centering
\includegraphics[width=\linewidth]{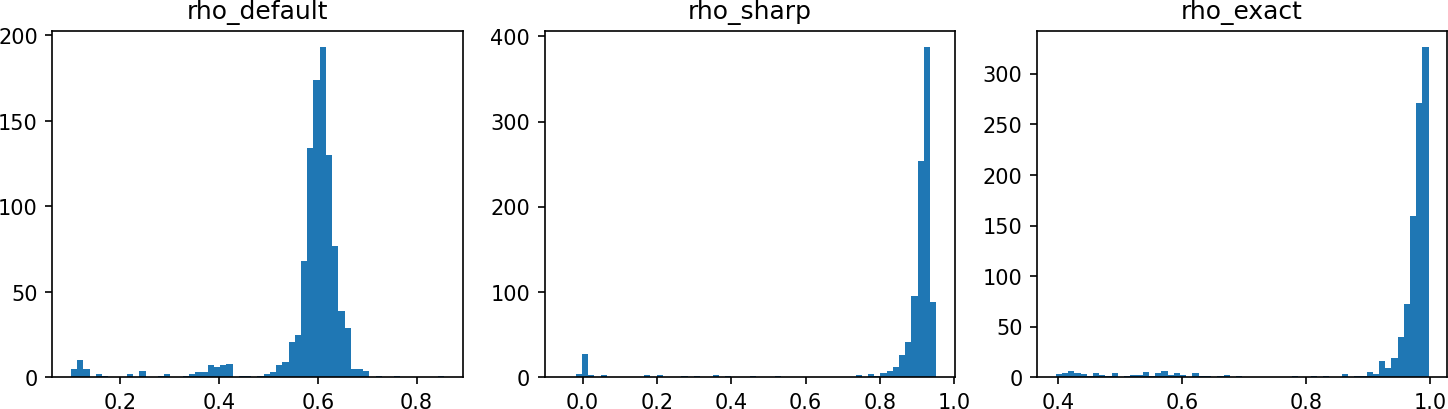}
\caption{Plan-alignment histograms. The separated low-alignment
group appears under sharp and hard plans. The $\rho^*$ notation in the
histogram refers to this statistic; only exact unregularized optimization
maximizes alignment over the sampled coupling set.}
\label{fig:app-rho}
\end{figure}
\FloatBarrier